\documentclass[nonacm]{acmart} 
\usepackage{amsmath}
\usepackage{adjustbox}
\usepackage{booktabs} 
\usepackage{color,soul}
\usepackage{graphicx}
\usepackage{diagbox}
\usepackage{makecell}
\usepackage{multirow,multicol}
\usepackage{textcomp}
\usepackage{setspace}
\usepackage{enumitem}
\usepackage{outlines}
\usepackage{tabularx}
\usepackage{makecell}
\usepackage{colortbl}
\usepackage{caption}
\usepackage{subcaption}  
\usepackage{algorithm2e}
\usepackage{array}
\usepackage{wrapfig}  

\AtBeginDocument{%
  }

\setcopyright{acmlicensed}
\copyrightyear{2018}
\acmYear{2018}
\acmDOI{XXXXXXX.XXXXXXX}
\acmConference[Conference acronym 'XX]{Make sure to enter the correct
  conference title from your rights confirmation email}{June 03--05,
  2018}{Woodstock, NY}
\acmISBN{978-1-4503-XXXX-X/2018/06}

\begin{document}

\title[CALLM]{CALLM: Understanding Cancer Survivors' Emotions and Intervention Opportunities via Mobile Diaries and Context-Aware Language Models}


\author{Zhiyuan Wang}
\affiliation{%
  \institution{Department of Systems and Information Engineering, University of Virginia}
  \city{Charlottesville}
  \state{Virginia}
  \country{United States}
}
\email{vmf9pr@virginia.edu}

\author{Katharine E. Daniel}
\affiliation{%
  \institution{Center for Behavioral Health and Technology, University of Virginia}
  \city{Charlottesville}
  \state{Virginia}
  \country{United States}
}
\email{ked4fd@virginia.edu}

\author{Laura E. Barnes}
\affiliation{%
  \institution{Department of Systems and Information Engineering, University of Virginia}
  \city{Charlottesville}
  \state{Virginia}
  \country{United States}
}
\email{lb3dp@virginia.edu}

\author{Philip I. Chow}
\affiliation{%
  \institution{Center for Behavioral Health and Technology, University of Virginia}
  \city{Charlottesville}
  \state{Virginia}
  \country{United States}
}
\email{pic2u@virginia.edu}

\renewcommand{\shortauthors}{Wang et al.}


\begin{abstract}
  Cancer survivors face unique emotional challenges that impact their quality of life. Mobile diary entries—short text entries people record through their personal phone—provide a promising method for tracking emotional states, improving self-awareness, and promoting well-being outcome. This paper aims to, through mobile diaries, understand cancer survivors' emotional states and key variables related to just-in-time intervention opportunities, including the desire to regulate emotions and the availability to engage in interventions. Although emotion analysis tools show potential for recognizing emotions from text, current methods lack the contextual understanding necessary to interpret brief mobile diary narratives. Our analysis of diary entries from cancer survivors (N=407) reveals systematic relationships between described contexts and emotional states—with administrative and health-related contexts associated with negative affect and regulation needs, while leisure activities promote positive emotions. We propose CALLM, a Context-Aware framework leveraging Large Language Models (LLMs) with Retrieval-Augmented Generation (RAG) to analyze these brief entries by integrating retrieved peer experiences and personal diary history. CALLM demonstrates strong performance with balanced accuracies reaching 72.96\% for positive affect, 73.29\% for negative affect, 73.72\% for emotion regulation desire, and 60.09\% for intervention availability, outperforming language model baselines. Post-hoc analysis reveals that model confidence strongly predicts accuracy, with longer diary entries generally enhancing performance, and brief personalization periods yielding meaningful improvements. Our findings demonstrate how contextual information in mobile diaries can be effectively leveraged to understand emotional experiences, predict key states, and identify optimal intervention moments for personalized just-in-time support.
\end{abstract}



\maketitle

\section{Introduction} \label{sec:intro}

In the context of cancer survivorship, understanding day-to-day emotional and behavioral fluctuations is critical to deliver tailored support and interventions. The National Cancer Institute (NCI) defines a cancer survivor broadly, encompassing individuals from diagnosis through the remainder of life \cite{mayer2017defining}. This large and growing population often faces significant and unique emotional challenges \cite{zebrack2000cancer, firkins2020quality}, including heightened psychological distress \cite{andrykowski2008psychological, powers2016treatment}, particularly in the initial years following diagnosis \cite{zabora2001prevalence}. Addressing these emotional needs is a key component of comprehensive survivorship care, linked to long-term well-being and recovery \cite{jacobs2017follow, salsman2023ehealth}. Despite the recognized importance of emotional support, many cancer survivors face barriers to accessing traditional mental health services, creating a need for accessible, low-burden approaches to emotional monitoring and intervention delivery \cite{hewitt2002mental,clifford2018barriers}.

Mobile diaries (e.g., brief text entries like ``Pain in my legs'' or ``Visit from my daughter'' in response to a prompt question ``what impacted your mood?''), allow users to submit brief open-ended journal entries through smartphone interfaces, representing a growing modality in mobile health and ubiquitous computing \cite{klasnja2012healthcare}. Mobile diaries can be collected through ecological momentary assessment (EMA), which captures users' experiences and states longitudinally, in-situ, as they occur in daily life via mobile phones \cite{shiffman2008ecological,de2021smartphone}. Crucially, these tools capture not just the \textit{what} (emotional state) but potentially the \textit{why} (the described context or `emotion driver'), providing rich contextual cues that can inform personalized interventions. If the qualitative data within these diaries can be effectively interpreted, it opens possibilities for numerous applications, including providing enhanced self-awareness \cite{morris2010mobile}, personalizing coaching feedback \cite{10.1145/3699723}, identifying moments of risk \cite{kleiman2018digital,wang2023power}, and tailoring adaptive interventions \cite{businelle2020reducing}.

For digital mental health interventions (DMHIs) to be effective, particularly within frameworks like just-in-time adaptive interventions (JITAIs) \cite{nahum2018just}, identifying and characterizing optimal ``intervention opportunities'' is crucial. Emotional states—like elevated negative affect or reduced positive affect—are well-established indicators of when psychological support may be most needed. In addition to emotional states, two other factors can further inform the timing and relevance of intervention delivery: (1) the subjective \textbf{desire} to regulate emotions—when individuals actively want to change their current emotional state, and (2) objective intervention \textbf{availability}—when individuals are practically able to engage with a mobile intervention. This dual requirement exists because even when individuals are available (not driving, in meeting, etc.) \cite{nahum2018just}, without the desire to change their emotions, motivation to engage will be minimal—a challenge common in conditions like depression where emotional inertia may limit regulation attempts \cite{tamir2016people}, despite distress. Conversely, strong motivation to regulate emotions is ineffective if practical constraints prevent intervention engagement. Understanding when either or both of these conditions are present can support more responsive and contextually appropriate intervention delivery \cite{hardeman2019systematic}. Analyzing the emotion drivers captured in mobile diaries potentially offers a low-burden way to characterize and detect these intervention opportunity cues without requiring multiple separate assessments.

However, unlocking the potential of mobile diaries for triggering and tailoring JITAIs hinges on accurately interpreting the states and their surrounding context reflected within the brief, often unstructured text entries. While computational techniques for analyzing health-related text show promise \cite{seabrook2018predicting,teodorescu2023language}, mobile diary data presents unique challenges that limit the direct application of methods developed for other text types (e.g., social media \cite{guntuku2019understanding, eichstaedt2018facebook}, clinical notes \cite{zeng2006audio}). Diary entries are characterized by their extreme brevity (in our dataset presented later, averaging only 6.83 words with a median word count of 4.0, as shown in Figure \ref{fig:length}), are subjective nature,  possess an introspective focus \cite{bolger2003diary}, and rely on implicit context. Furthermore, entries often lack explicit emotional language (66.2\% exhibited neutral sentiment polarity in our data, as shown in Figure \ref{fig:sentiment_neutral}), making state inference difficult. Traditional text analysis methods, whether based on feature engineering or standard deep learning \cite{wang2016dimensional}, often falter with such sparse data. Even recent LLMs, while powerful, typically require explicit prompting or fine-tuning for specific tasks and may struggle to capture the necessary personal and temporal context when processing entries in isolation \cite{xu2024mental, stone2020evaluating, 10.1145/3173574.3173987}. There is a need for methods that can effectively leverage the limited text alongside personal contexts and trajectories to understand the emotional and behavioral states reflected in mobile diaries.

In this paper, our investigation is guided by the following research questions:
\begin{itemize}[topsep=2pt, partopsep=2pt, itemsep=0pt, parsep=0pt]
    \item[\textbf{RQ1:}] What contextual information (e.g., described activities, social interactions, health factors) is conveyed in brief mobile diary entries from cancer survivors, and how does this described context relate to their concurrent emotional states, desire to regulate emotions, and availability to engage in interventions?
    \item[\textbf{RQ2:}] Can individuals' emotional states, desire to regulate emotions, and availability to engage in interventions be inferred from brief mobile diary text using a context-aware LLM framework?
    \item[\textbf{RQ3:}] How do factors such as model confidence, diary entry length, the inclusion of personal temporal context, and personalization using initial user data influence the performance of LLM-based prediction of emotional states, desire to regulate emotions, and availability to engage in interventions from mobile diaries?
\end{itemize}

To address these research questions, we conducted a study involving the collection and analysis of a large-scale, longitudinal mobile diary dataset. First, to enable the contextual analysis required for RQ1, we collected and characterized 24,183 diary entries over five weeks from \(N=407\) cancer survivors, capturing their emotional experiences alongside self-reported states, as well as interpreting them with contextual cues captured from the diaries. Second, to investigate RQ2—predicting emotional states, desire to regulate emotions, and availability to do interventions from these brief entries—we developed and evaluated CALLM (Context-Aware language model framework for mobile diaries). This framework leverages LLMs augmented with context dynamically retrieved from peer experiences (via retrieval augmentation; RAG \cite{zhao2024retrieval}) and the individual's own temporal trajectory from diary history. Third, to explore the factors influencing performance (RQ3), we compared CALLM against various baselines and conducted extensive post-hoc analyses examining the impact of LLMs' confidence, diary entry length, diaries' temporal dependency, and personalization potential.

This work offers the following contributions:
\begin{itemize}[topsep=2pt, partopsep=2pt, itemsep=0pt, parsep=0pt]
    \item[1.] We collected and analyzed a large-scale mobile diary dataset from cancer survivors, revealing systematic relationships between described contexts and emotional states. This characterization provides insights into how different daily activities and contexts influence cancer survivors' emotional experiences and identifies patterns relevant for contextually-aware intervention delivery.
    \item[2.] We designed and evaluated CALLM, a novel context-aware framework that enhances LLM-based analysis of brief mobile diary entries by integrating peer experiences and individual temporal trajectories. Our approach demonstrates how contextual enrichment can significantly improve the interpretation of sparse text data for understanding intervention opportunities by detecting cancer survivors' self-reported emotional states and desire to regulate emotions, though availability detection still leaves room for improvement.
    \item[3.] We identified key factors that influence prediction performance through comprehensive post-hoc analyses, exploring how model confidence, entry length, temporal context, and personalization affect accuracy. These findings provide practical guidance for implementing mobile diary analysis systems in real-world settings.
\end{itemize}

We envision this work informing the development of context-aware JITAI systems that leverage mobile data to deliver highly personalized and precisely timed healthcare interventions. By interpreting user-generated entries—and potentially integrating passive sensing data from smartphones and wearables in future studies—this approach supports interventions attuned to individuals’ emotional states, desire, and availability, aligning closely with their real-world motivations and contexts. Ultimately, this direction advances adaptive, user-centered digital health systems capable of supporting individuals through complex emotional and behavioral journeys.

\section{Related Work}

\subsection{Mobile Diaries for In-Situ Health Monitoring} \label{sec:mobile-diaries}

Mobile diaries, often implemented through smartphone applications as part of ecological momentary assessment (EMA) protocols, are increasingly utilized in mobile health research. Their primary strength lies in capturing data longitudinally and \textit{in-situ}, reflecting experiences and states as they naturally occur in daily life \cite{shiffman2008ecological}. This approach offers greater ecological validity and minimizes recall bias compared to retrospective self-reports \cite{stinson2022ecological}. Mobile diaries, particularly free-text formats, can achieve this with relatively lower subjective burden than frequent, lengthy structured questionnaires \cite{mustafa2022user}. Common applications include monitoring symptoms in chronic illnesses \cite{krieke2016hownutsarethedutch}, tracking mood and behavior patterns in mental health contexts \cite{davies2023individual}, and understanding daily experiences in specific populations like cancer survivors \cite{funk2020framework, metsaranta2019adolescents}. However, realizing the full potential of this rich qualitative data is often hindered by challenges including participant compliance over time, the inherent subjectivity of entries, and, most relevantly for computational analysis, the extreme brevity and sparsity of typical free-text diary entries \cite{sohn2008diary,linton2021investigating,stone2023evaluation}. This necessitates advanced methods capable of interpreting meaning from limited text input. These challenges, particularly the need to extract meaning from sparse text while considering individual context, motivate the exploration of context-aware systems (Section \ref{sec:context-aware}) and specialized text analysis techniques (Section \ref{sec:analyzing-brief}).

\subsection{Context-Aware Systems in Mobile Health} \label{sec:context-aware}

Understanding user context is a central theme in ubiquitous computing and mobile health \cite{abowd1999towards,harari2023understanding}. Context-aware systems aim to adapt their behavior based on situational information—such as location, activity, social setting, and human behavioral or emotional states—to provide more relevant and effective support \cite{10.1145/3448080,10.1145/3432213,10.1145/3580798}. In mobile health, context plays a critical role in understanding health behaviors and enabling timely interventions, such as JITAIs, where algorithms are designed to adapt dynamically to changing contextual factors \cite{nahum2018just,10.1145/3287031}. Previous work has also explored leveraging diverse signals for understanding user's general contexts or contexts relevant to their disorders, including integrating passive sensor data (e.g., GPS, accelerometer, phone usage logs, heart rate, and audio signals) \cite{mohr2017personal, straczkiewicz2021systematic,wang2023detecting,reddy2024audioinsight} and employing multi-modal approaches that combine text with audio or visual cues \cite{poria2019emotion}. Complementary to these approaches, our work focuses on the challenge of extracting rich contextual understanding directly \textit{from the user's own words} within brief, free-text mobile diary entries, potentially offering a low-friction method for capturing subjective context often missed by sensors alone. In particular, we characterize how cancer survivors as a unique population experience emotional deviations across different contextual situations (e.g., administrative, health-related, leisure activities) and express intervention opportunities through their diary narratives, enabling the identification of context-specific moments for targeted support without requiring additional sensing infrastructure.

\begin{figure}[t!]
    \centering
    \begin{subfigure}{0.47\textwidth}
        \includegraphics[width=\textwidth]{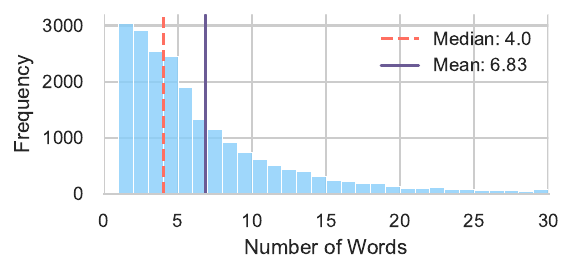}
        \caption{Distribution of word counts in non-empty diary entries.}
        \label{fig:length}
    \end{subfigure}
    \hfill
    \begin{subfigure}{0.47\textwidth}
        \includegraphics[width=\textwidth]{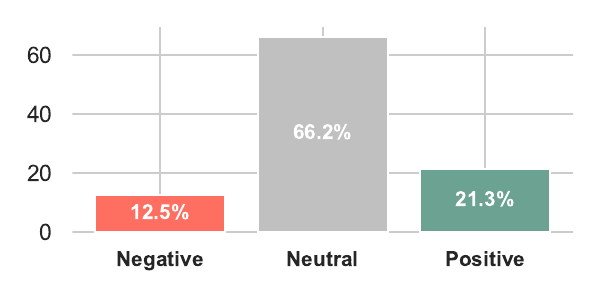}
        \caption{Distrubution of diary sentiment polarity based on TextBlob.}
        \label{fig:sentiment_neutral}
    \end{subfigure}
    \caption{Distrubutions of word count and sentiment polarity across collected mobile diaries from N=407 cancer survivors.}
    \label{fig:challenges}
\end{figure}

\subsection{Analyzing Mobile Diary Text with Language Models} \label{sec:analyzing-brief}

Traditional emotion analysis has primarily targeted content-rich texts with explicit emotional cues, such as clinical interviews and social media posts (averaging $420$ tokens\footnote{In language models, a token is the smallest unit of text produced by segmentation and is not necessarily a complete word. Typically, one word equals about 1.33 tokens, though this ratio can vary by tokenizer and model.} per post in Reddit datasets \cite{turcan2019dreaddit}), where contextual richness supports mental health assessments \cite{seabrook2018predicting,teodorescu2023language}. These approaches leverage deep learning models \cite{wang2016dimensional}, domain-specific frameworks \cite{funk2020framework}, and increasingly, LLMs \cite{xu2024mental}, thriving in contexts where users curate rich, expressive content \cite{papacharissi2010networked}. However, mobile diary entries present fundamentally different challenges with their extreme brevity—averaging merely $7.57$ tokens ($6.83$ words) per entry in this study's dataset (Figure \ref{fig:length})—and introspective nature. Unlike public-facing texts, these private reflections \cite{bolger2003diary} often lack explicit emotional language ($66.2\%$ of entries demonstrate neutral sentiment polarity in our dataset, Figure \ref{fig:sentiment_neutral}) and contextual redundancy. Existing machine learning methods struggle with this textual sparsity due to: (1) dependency on predetermined task designs unsuited for minimal text \cite{guntuku2019understanding,eichstaedt2018facebook}, (2) inflexibility in adapting to personalized emotional trajectories \cite{10.1145/3173574.3173987}, and (3) insufficient contextual awareness for interpreting ultra-brief entries \cite{stone2020evaluating}.

Recent advances in LLMs have shown remarkable capabilities across various domains \cite{yang2022large}, particularly in mental health applications. These include detecting cognitive distortions \cite{xu2024mental,chen2023empowering}, analyzing depression through clinical interview data \cite{sadeghi2023exploring}, supporting cognitive reframing of negative thoughts \cite{sharma2023cognitive}, and facilitating empathic peer-to-peer conversations in therapeutic settings \cite{sharma2023human, wang2024rapport, wang2024pallm}. RAG techniques further enhance LLMs by incorporating external knowledge, as demonstrated in personalized abusive language detection \cite{yao2024personalised} and other knowledge-intensive tasks \cite{lewis2020retrieval}, while interpretable mental health analysis approaches offer explainable predictions \cite{yang2023towards} and well-being support \cite{ma2024understanding}. In the specific domain of digital journaling, emerging work has begun exploring LLM integration: Jung et al.'s \emph{MyListener} combines smartphone and Fitbit data to alleviate depression and loneliness \cite{10.1145/3675094.3677601}, Kim et al.'s \emph{MindfulDiary} improves journaling consistency through psychiatric-informed conversational interfaces \cite{10.1145/3613904.3642937}, and Nepal et al. \cite{10.1145/3613905.3650767,10.1145/3699761} generates contextual journaling prompts from behavioral sensing data. While these approaches demonstrate LLMs' potential for mental health applications, they primarily focus on enhancing the journaling experience rather than deriving rich insights from minimal text entries.

Building on existing work and addressing key challenges, the CALLM framework uniquely integrates peer experiences and personal temporal trajectories—both derived directly from diary entries—to extract meaningful insights from sparse text. It shows how even brief entries can illuminate cancer survivors’ life contexts and emotional needs, thereby enabling new applications of mobile diaries.

\section{Methods}

\subsection{Participants}
The present study involved data gathered from \textit{N} = 407 US adults within five years of a cancer diagnosis (stages 0-4, referred to as `cancer survivors') and who owned a smartphone. Participants were eligible regardless of cancer type and treatment status. Of the 426 who were enrolled, N = 19 were excluded (n = 10 did not initiate the EMA, n = 9 requested to withdraw), resulting in a final sample of N = 407. Participants were, on average, 48.73 years old (SD = 12.23); 9.09\% male and 90.17\% female; 86.67\% White; and 92.52\% non-Hispanic. Most participants (57.28\%) had a primary diagnosis of breast cancer and 21.88\% were actively receiving cancer treatment during the study. The study protocol has been fully approved by the Institutional Review Board at the University of Virginia IRB \#HSR230080.

\subsection{Procedure}
\subsubsection{Recruitment Strategy}
Participants were recruited via targeted online advertising. Recruitment was promoted by BuildClinical\footnote{\url{https://www.buildclinical.com/}}, which specializes in supporting robust and diverse recruitment for academic research trials. BuildClinical deploys advertisements according to data-driven strategy to target and engage specific patient populations through their large digital network (e.g., patient communities, Google, Facebook, Instagram, health websites, medical apps). Their advertising campaigns engage potentially eligible participants by deploying custom advertisements across relevant platforms using applied machine-learning and data mining to identify digital footprints of specific patient groups. During the study, algorithms are refined to engage specific groups and patient populations of interest. 

\subsubsection{Enrollment Process}
Interested individuals responded to an online advertisement and completed an online pre-screen survey to verify eligibility (i.e., diagnosed with cancer in the last 5 years and owns a smartphone). Pre-eligible individuals had to pass a background check to verify their identity and the consent process was conducted on the phone. Prior to initiating the EMA phase, participants completed a battery of self-report questionnaires to assess demographics, cancer diagnosis and treatment history, and psychological functioning.

\subsubsection{EMA Protocol}
During the five-week EMA phase, participants received three surveys per day via the Effortless Assessment Research System (EARS) mobile application. Surveys were delivered randomly within three 2-hour time windows: morning (8-10am), afternoon (1-3pm), and evening (7-9pm). This schedule was designed to capture how psychological, behavioral, and emotional processes unfold throughout the day. Participants were notified of new surveys via push notifications, and surveys expired if not answered within 2 hours.

\begin{figure}[t!]
    \centering
    \begin{subfigure}{0.23\textwidth}
        \includegraphics[width=\textwidth]{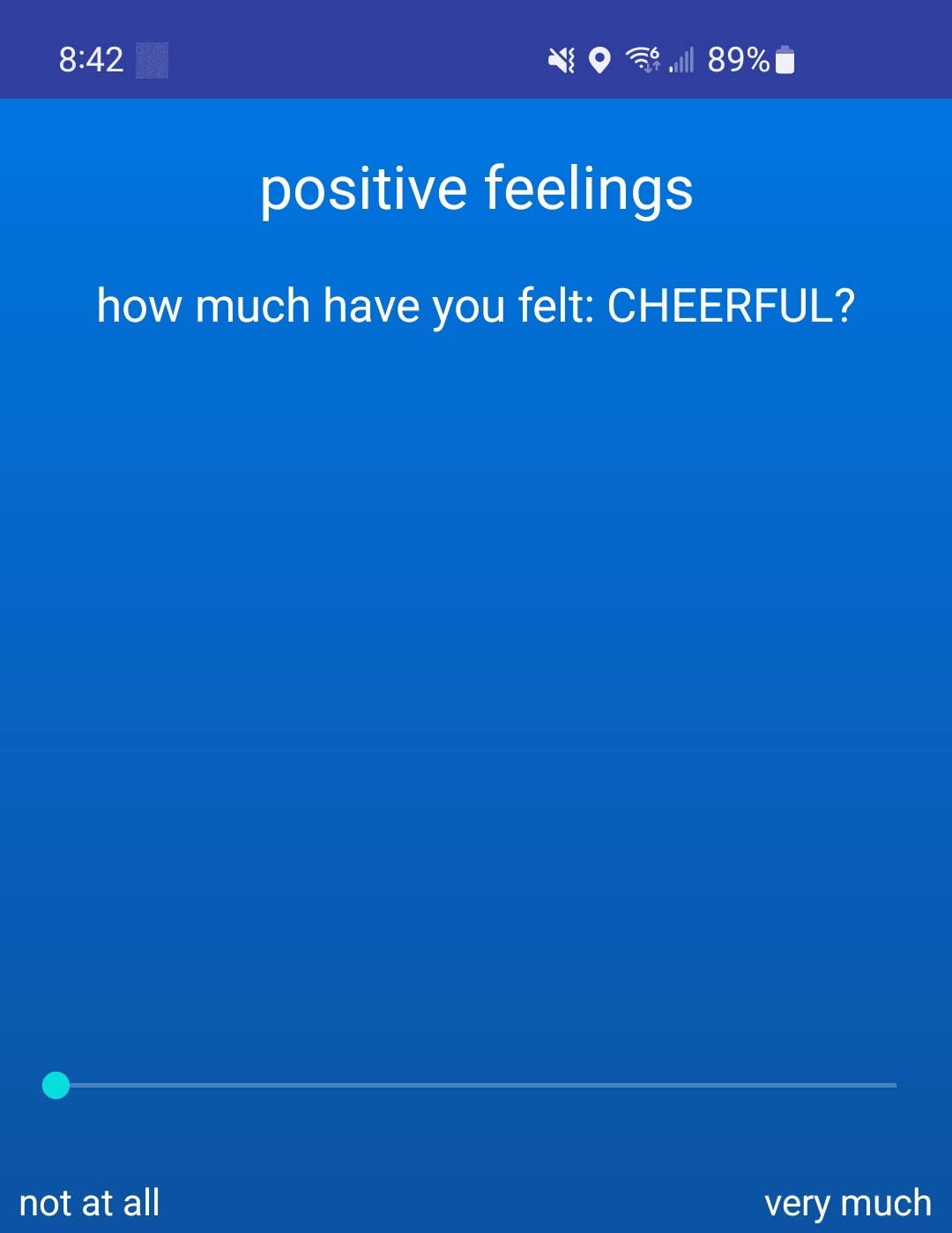}
        \caption{Positive emotions.}
        \label{fig:EMA_1}
    \end{subfigure}
    \begin{subfigure}{0.23\textwidth}
        \includegraphics[width=\textwidth]{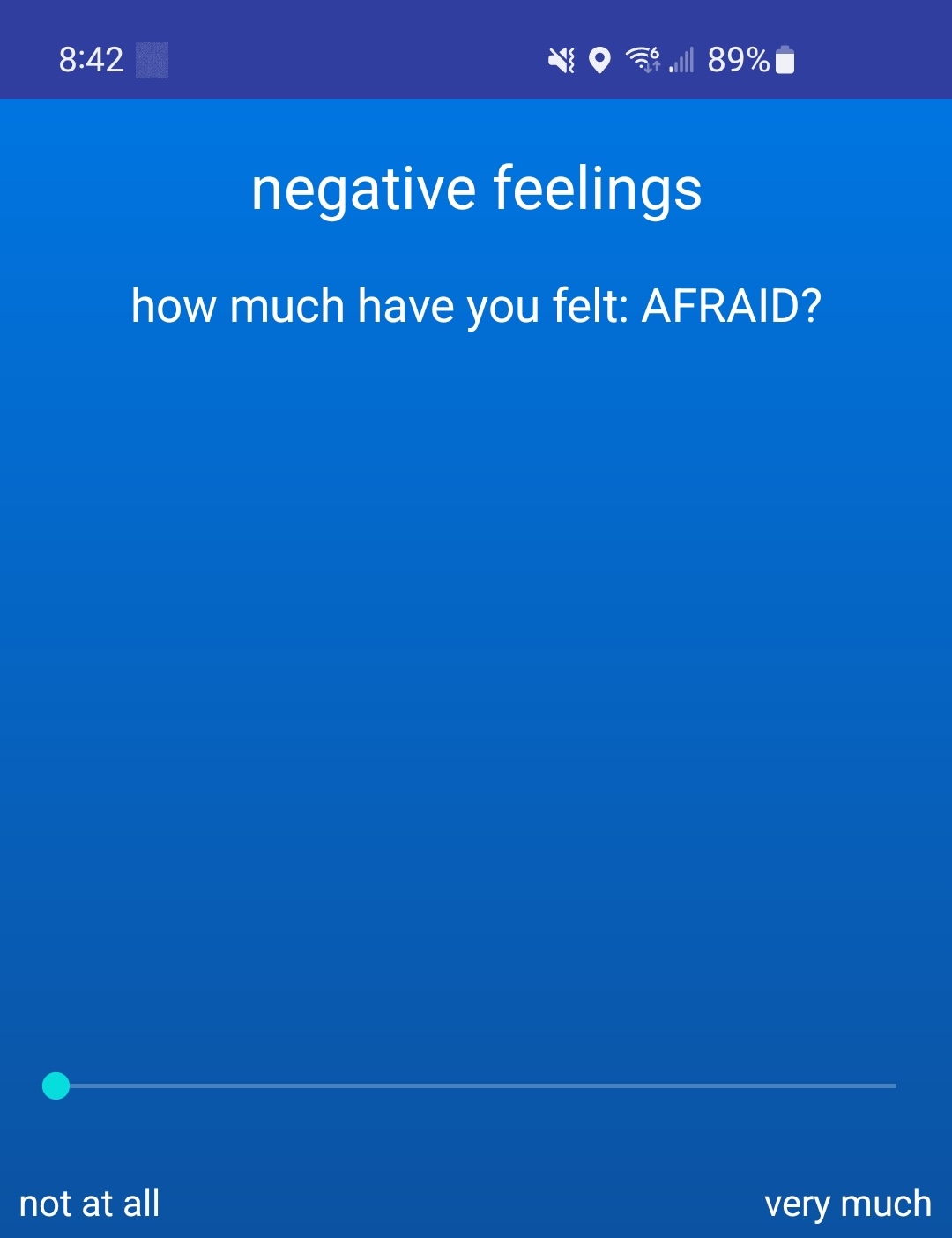}
        \caption{Negative emotions.}
        \label{fig:EMA_2}
    \end{subfigure}
    \begin{subfigure}{0.228\textwidth}
        \includegraphics[width=\textwidth]{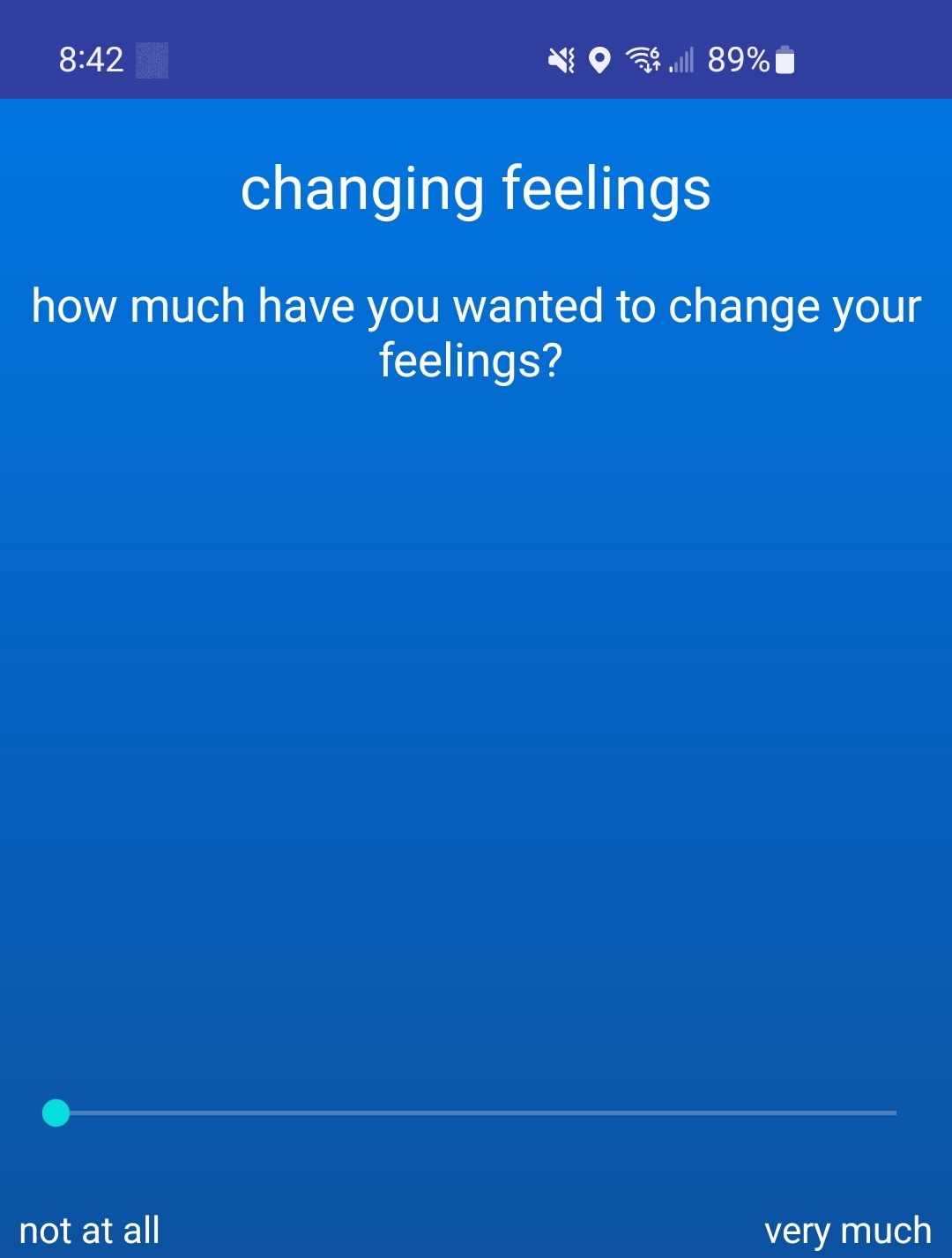}
        \caption{Desire to regulate emotions.}
        \label{fig:EMA_3}
    \end{subfigure}
    \begin{subfigure}{0.23\textwidth}
        \includegraphics[width=\textwidth]{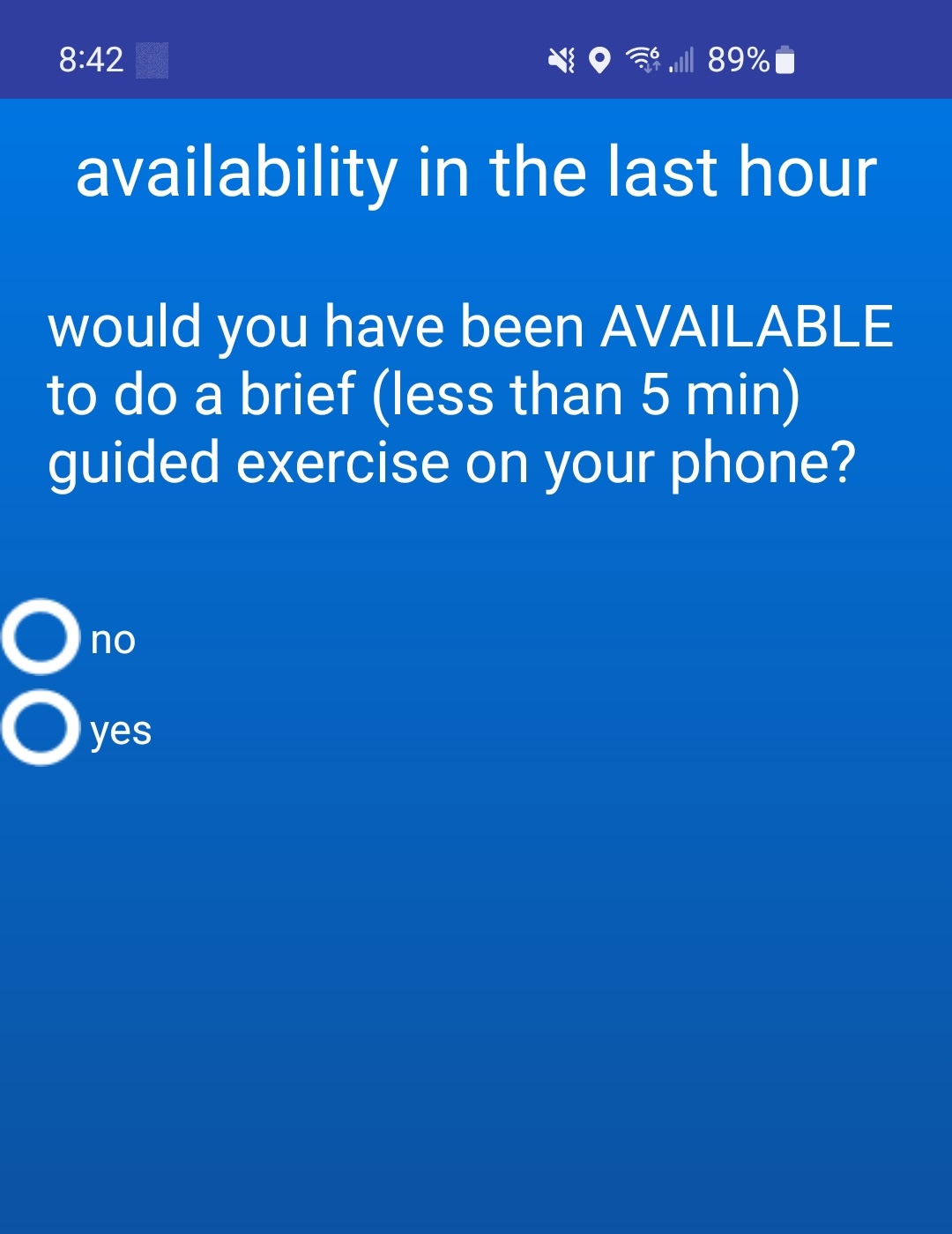}
        \caption{Intervention availability.}
        \label{fig:EMA_4}
    \end{subfigure}
    \caption{EMA interface for collecting emotional measures, desire to regulate emotions, and availability to do interventions.}
    \label{fig:EMAs}
\end{figure}

\subsubsection{Compensation Structure}
Participants received up to \$100 in gift cards for completing the study, with prorated compensation:
\begin{itemize}
    \item \$20 for completing the baseline questionnaire battery
    \item Additional \$50 for completing between 50\%-75\% of the EMA surveys
    \item Or, additional \$80 for completing at least 75\% of the EMA surveys 
\end{itemize}

\subsection{EMA Measures}
\subsubsection{Primary Outcome Measures}
\begin{itemize}
    \item \textbf{Emotion driver diary entry:} Participants responded to the prompt regarding the last hour, ``What has had the biggest impact on your mood?'' using an open text box on their smartphone.

    \item \textbf{State affect:} Participants rated the degree to which they felt specific emotions in the past hour using a 0 (\textit{not at all}) to 10 (\textit{very much}) scale:
    \begin{itemize}
        \item \textbf{Positive Affect:} Sum of ratings for \texttt{happy}, \texttt{cheerful}, and \texttt{pleased} (range: 0-30)
        \item \textbf{Negative Affect:} Sum of ratings for \texttt{sad}, \texttt{afraid}, and \texttt{miserable} (range: 0-30)
    \end{itemize}

    \item \textbf{Desire to regulate emotions:} Participants were asked regarding the last hour, ``How much have you wanted to change your feelings?'' using a 0 (\textit{not at all}) to 10 (\textit{very much}) scale.

    \item \textbf{Intervention availability:} Participants were asked regarding the last hour, ``Would you have been AVAILABLE to do a brief (less than 5 minute) guided exercises on your phone?'' using a yes/no response format.
\end{itemize}

\subsubsection{Secondary Outcome Measures}
Additional variables were measured using a 0 (\textit{not at all}) to 10 (\textit{very much}) Likert scale: \texttt{social interaction quality}, \texttt{pain}, \texttt{worried}, \texttt{lonely}, and \texttt{grateful}.

\subsubsection{Binary State Indicators}
To standardize analysis across participants and account for individual differences in baseline levels and response biases, we created binary state indicators for each emotional and behavioral measure. These indicators reflect whether a participant's momentary rating exceeds their own mean level (True) or not (False) for that measure, allowing us to identify periods of elevated states relative to each participant's typical experience.

\section{Characterizing Contextual Information in Mobile Diaries} \label{sec:contextual_info}

To answer \textbf{RQ1} (concerning contextual information in mobile diary entries and its relationship to emotional states), we systematically analyzed 24,183 diary entries collected from \(N=407\) cancer survivors. Our analysis approach was multi-faceted, examining (1) the relationship between diary topic categories and emotional/behavioral states, (2) how sentiment expressed in diary entries relates to these states, and (3) lexical patterns characterizing high-intensity emotional experiences. These analyses collectively provide insights into what contextual cues are present in brief diary entries and how they might inform the triggering and personalization of mobile interventions.

\subsection{Topic-Based Analysis of Emotional States} \label{sec:topic_analysis}

\begin{figure}[t!]
    \centering
    \begin{subfigure}{0.49\textwidth}
        \includegraphics[width=\textwidth]{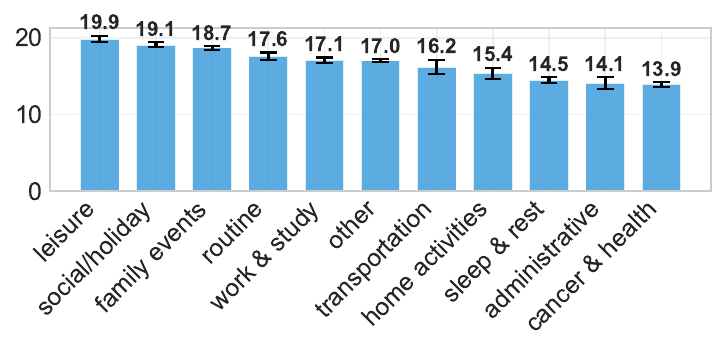}
        \caption{Positive Affect (0-30)}
        \label{fig:topic_analysis_a}
    \end{subfigure}
    \hfill
    \begin{subfigure}{0.49\textwidth}
        \includegraphics[width=\textwidth]{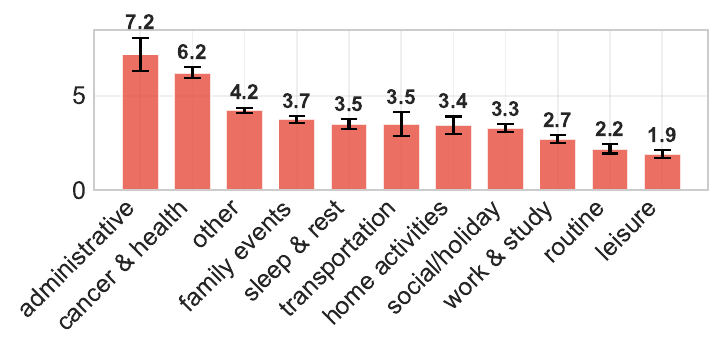}
        \caption{Negative Affect (0-30)}
        \label{fig:topic_analysis_b}
    \end{subfigure}
    
    \vspace{4mm}
    \begin{subfigure}{0.49\textwidth}
        \includegraphics[width=\textwidth]{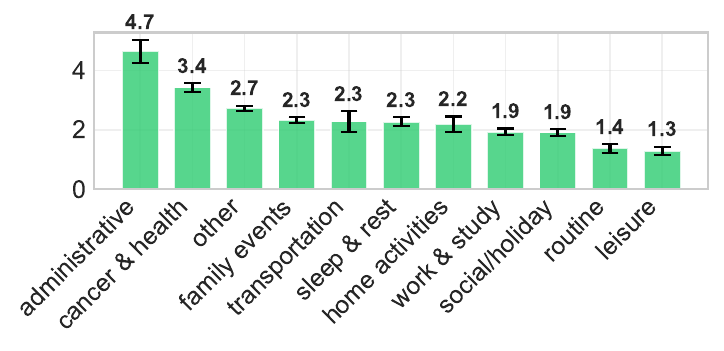}
        \caption{Emotion Regulation Desire (0-10)}
        \label{fig:topic_analysis_c}
    \end{subfigure}
    \hfill
    \begin{subfigure}{0.49\textwidth}
        \includegraphics[width=\textwidth]{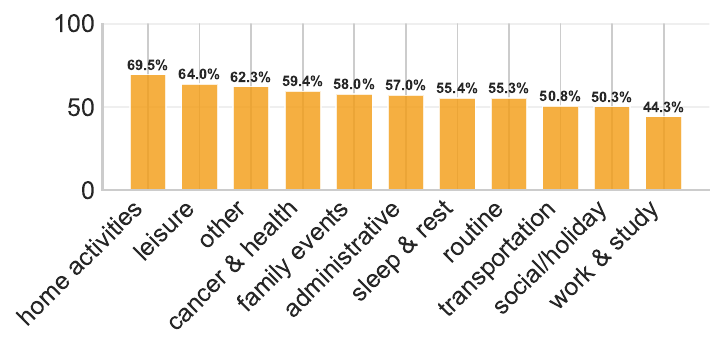}
        \caption{Intervention Availability (Yes or No)}
        \label{fig:topic_analysis_d}
    \end{subfigure}
    
    \caption{Emotional and Behavioral Measures Across Different Emotion Driver Categories.}
    \label{fig:topic_analysis}
\end{figure}
To analyze how different activity contexts relate to emotional and behavioral measures, we categorized diary entries into context groups through keyword matching. The groups include: administrative, cancer \& health, family events, home activities, leisure, outdoor activity, pet activities, routine, sleep \& rest, social/holiday, transportation, work \& study, and other. Each entry could be assigned to multiple groups based on its content (e.g., ``walked dog in park'' would be categorized under both ``outdoor activity'' and ``pet activities''). The `Others' category includes diaries that are too brief or lack the necessary keywords to provide sufficient information for topic matching.

Analysis of emotional and behavioral measures across different context categories reveals distinct patterns in how various activities and contexts relate to cancer survivors' emotional experiences (Figure~\ref{fig:topic_analysis}). For positive affect (Figure~\ref{fig:topic_analysis}a), leisure activities (mean=19.90), social/holiday events (mean=19.11), and family events (mean=18.73) demonstrate notably higher scores compared to administrative tasks (mean=16.18) or cancer/health-related activities (mean=15.87). Mixed effects models, accounting for participant-level clustering (which explained 61.6\% of variance), confirmed these differences were statistically significant. Compared to administrative tasks (reference category), significantly higher scores were observed for contexts related to leisure activities (coef=3.97, adjusted $p<0.001$), social/holiday events (coef=4.16, adjusted $p<0.001$), and family events (coef=4.39, adjusted $p<0.001$).

Negative affect distributions (Figure~\ref{fig:topic_analysis}b) show an inverse pattern, with cancer/health-related categories (mean=6.16) and administrative tasks (mean=5.72) exhibiting higher mean scores and greater variability, particularly evident in the numerous outliers. Work and study activities show moderate negative affect levels (mean=2.71), highlighting the emotional challenges cancer survivors face in managing daily responsibilities. After controlling for individual differences (which explained 59.0\% of variance), significantly lower negative affect scores were found for leisure (coef=-3.23, adjusted $p<0.001$), cancer/health-related categories (coef=-3.63, adjusted $p<0.001$), social/holiday events (coef=-2.94, adjusted $p<0.001$), and work \& study (coef=-2.90, adjusted $p<0.001$) compared to administrative tasks.

Emotion regulation desire (Figure~\ref{fig:topic_analysis}c) peaks during administrative tasks (mean=4.44) and cancer/health-related activities (mean=3.50), while being notably lower during leisure (mean=1.29), outdoor activities (mean=1.34), and routine activities (mean=1.37). Statistical analysis, controlling for individual differences (which explained 48.9\% of variance), confirmed that all categories showed significantly lower scores compared to administrative tasks: cancer \& health (coef=-0.94, adjusted $p<0.001$), leisure (coef=-2.65, adjusted $p<0.001$), outdoor activities (coef=-2.53, adjusted $p<0.001$), social/holiday events (coef=-2.44, adjusted $p<0.001$), and work \& study (coef=-2.02, adjusted $p<0.001$).

The availability to engage with digital interventions (Figure~\ref{fig:topic_analysis}d) varies considerably across contextual categories, with home activities showing the highest availability (69.5\%), while work and study contexts demonstrate the lowest (44.3\%). Chi-square tests confirmed that these differences in intervention availability across topic categories were statistically significant ($\chi^2(12) = 313.5$, adjusted $p < 0.001$, Cramer's $V = 0.12$), indicating a small to medium association between activity context and receptiveness to digital intervention. These findings have direct implications for the targeting of just-in-time interventions, suggesting that both emotional need and practical availability should be considered when determining optimal intervention moments.

\begin{figure}[t!]
    \centering
    \begin{subfigure}{0.65\textwidth}
        \includegraphics[width=\textwidth]{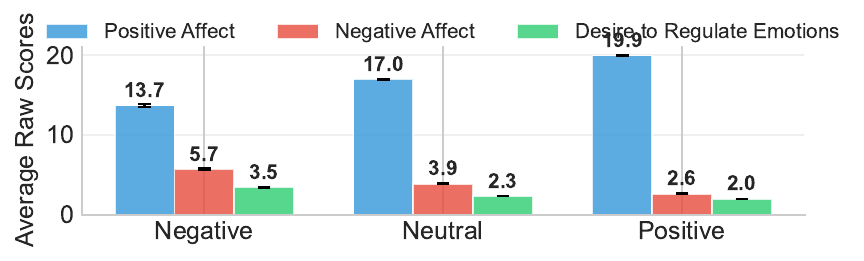}
        \caption{Self-reported Raw Scores (0-30 or 0-10)}
        \label{fig:sentiment-analysis_a}
    \end{subfigure}
    \hfill
    \begin{subfigure}{0.32\textwidth}
        \includegraphics[width=\textwidth]{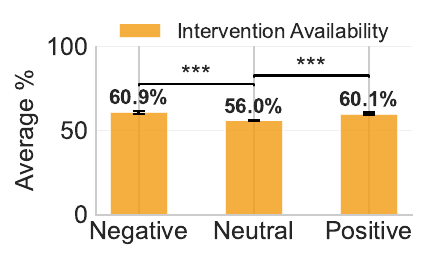}
        \caption{Intervention Availability (Yes/No)}
        \label{fig:sentiment-analysis_b}
    \end{subfigure}
    
    \caption{Emotional States and Intervention Availability Across Different Sentiment Categories}
    \label{fig:sentiment-analysis}
\end{figure}

\subsection{Sentiment Analysis and Emotional States} \label{sec:sentiment_analysis}

While topic analysis provides insights into how activity contexts relate to emotional states, the sentiment expressed within diary entries offers another dimension of contextual understanding. We performed sentiment analysis on diary entries using TextBlob toolkit \cite{loria2018textblob}, resulting in continuous sentiment scores ranging from -1.0 to 1.0. Entries were categorized as negative (score < -0.1, 12.5\%), neutral (score = -0.1 to 0.1, 66.2\%), or positive (score > 0.1, 21.3\%). The predominance of neutral sentiment entries (66.2\%) highlights the challenge of emotion inference from brief diary texts, as participants often describe situations factually without explicit emotional polarity.

Analysis of emotional self-reports by sentiment category confirms significant differences in experienced emotions that align with the detected text sentiment (Figure~\ref{fig:sentiment-analysis}a). For affective states, positive sentiment entries showed the highest positive affect scores (mean=19.95), followed by neutral (mean=16.98) and negative (mean=13.66) entries, while negative sentiment entries showed the highest negative affect (mean=5.71), followed by neutral (mean=3.86) and positive (mean=2.62) entries. Mann-Whitney U tests confirmed these differences were statistically significant for both positive affect (negative vs. neutral: $U=13.8M$, $p<0.001$, $r=0.24$; positive vs. neutral: $U=37.6M$, $p<0.001$, $r=-0.21$; negative vs. positive: $U=3.3M$, $p<0.001$, $r=0.44$) and negative affect (negative vs. neutral: $U=21.3M$, $p<0.001$, $r=-0.20$; positive vs. neutral: $U=26.2M$, $p<0.001$, $r=0.13$; negative vs. positive: $U=7.6M$, $p<0.001$, $r=-0.33$).

The desire to regulate emotions varied similarly across sentiment categories, with the highest scores in negative sentiment entries (mean=3.45), followed by neutral (mean=2.32) and positive (mean=1.98) entries, with Mann-Whitney U tests confirming statistical significance (negative vs. neutral: $U=22.0M$, $p<0.001$, $r=-0.21$; negative vs. positive: $U=7.6M$, $p<0.001$, $r=-0.29$). Interestingly, intervention availability (Figure~\ref{fig:sentiment-analysis}b) showed a \texttt{U}-shaped pattern, with higher availability during moments where diary entries exhibit any polarity, either negative (60.9\%) or positive (60.1\%), compared to neutral moments (56.0\%). Chi-square tests confirmed these differences were statistically significant ($\chi^2(2)=38.2$, $p<0.001$), though the effect size was small (Cramer's $V=0.04$). This finding suggests that participants may be more receptive to interventions during emotionally salient moments (either positive or negative) compared to neutral states.

\begin{figure}[t!]
    \centering
    \begin{subfigure}{0.48\textwidth}
        \includegraphics[width=\textwidth]{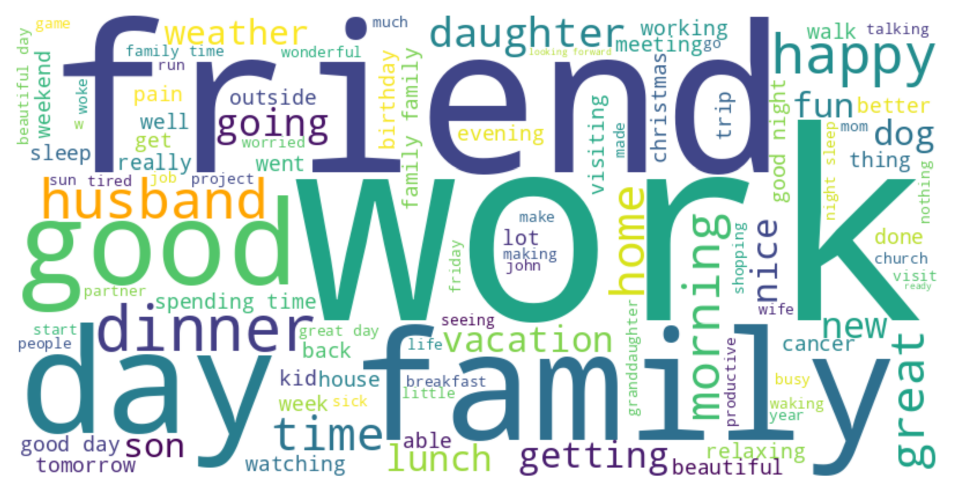}
        \caption{Positive Affect.}
        \label{fig:word_cloud-analysis_a}
    \end{subfigure}
    \hfill
    \begin{subfigure}{0.48\textwidth}
        \includegraphics[width=\textwidth]{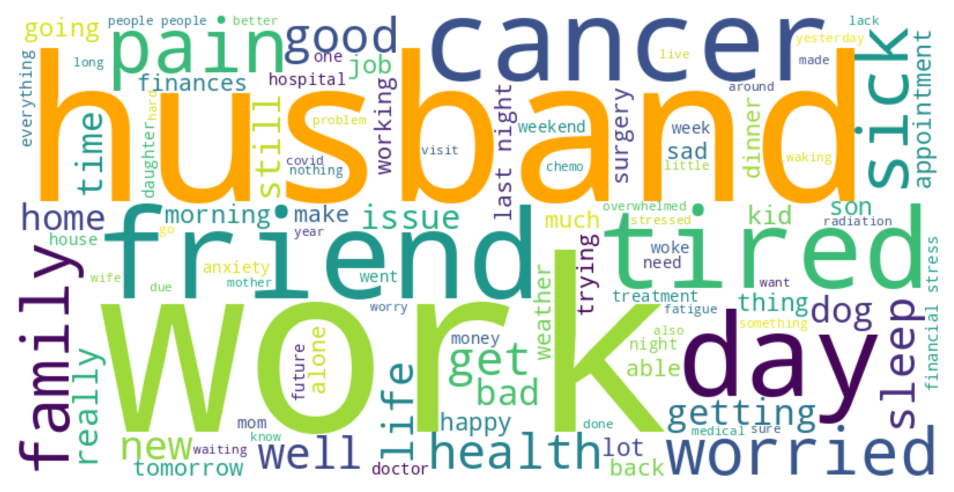}
        \caption{Negative Affect.}
        \label{fig:word_cloud-analysis_b}
    \end{subfigure}
    
    \caption{Word clouds with high (above individual mean) affective scores reported.}
    \label{fig:word_cloud}
\end{figure}

\subsection{Lexical Patterns in Emotional Experiences} \label{sec:word_cloud_analysis}

Beyond topics and sentiment, examining the specific vocabulary used in diary entries provides deeper insights into the contextual factors driving cancer survivors' emotional experiences. To visualize these patterns, we generated word clouds based on entries associated with high emotional intensity (defined as above individual mean plus one standard deviation for both positive and negative affect). Common English stop words (e.g., ``the'', ``is'', ``at'') and emotion-specific terms (e.g., ``feel'') were removed to focus on content-bearing words.

Analysis of the word clouds (Figure~\ref{fig:word_cloud}) reveals distinct vocabulary patterns between positive and negative affect contexts. In positive affect entries (Figure~\ref{fig:word_cloud}a), daily activities and social connections dominate the discourse: \texttt{work}, \texttt{day}, and \texttt{time} appear as central terms, while social terms like \texttt{family}, \texttt{daughter}, and \texttt{friend} feature prominently. Activity-related words such as \texttt{dinner}, \texttt{vacation}, and \texttt{walking} suggest the importance of both routine and recreational activities in positive experiences. 

The negative affect word cloud (Figure~\ref{fig:word_cloud}b) presents a markedly different emotional landscape. Health-related terms are most prominent, with \texttt{cancer} appearing as one of the largest words, accompanied by terms like \texttt{pain}, \texttt{tired}, and \texttt{health}. Relationship terms also feature differently here - while \texttt{family} remains present, \texttt{husband} appears with notable prominence (notably, 89.93\% of participants are reported as female, n=375). The presence of terms like \texttt{worried}, \texttt{anxiety}, and \texttt{stress} directly reflects emotional challenges, while words related to medical experiences (\texttt{hospital}, \texttt{treatment}, \texttt{doctor}) indicate the ongoing impact of health concerns on daily life.

\section{Predicting Emotional States and Intervention Opportunities from Mobile Diary Text}
To address \textbf{RQ2} (whether nuanced emotional states, desire and availability relevant to intervention opportunities can be accurately predicted from brief mobile diary text using a context-aware LLM framework), we developed and evaluated CALLM (Context-Aware Language Model). This section details both the design of our framework and its evaluation results, demonstrating how brief text entries can be leveraged to predict emotional states, desire to regulate emotions, and availability to engage in interventions for personalized intervention delivery.

\subsection{CALLM: A Context-Aware Framework for Mobile Diary Analysis}

\begin{figure*}
    \centering
    \includegraphics[width=0.9\linewidth]{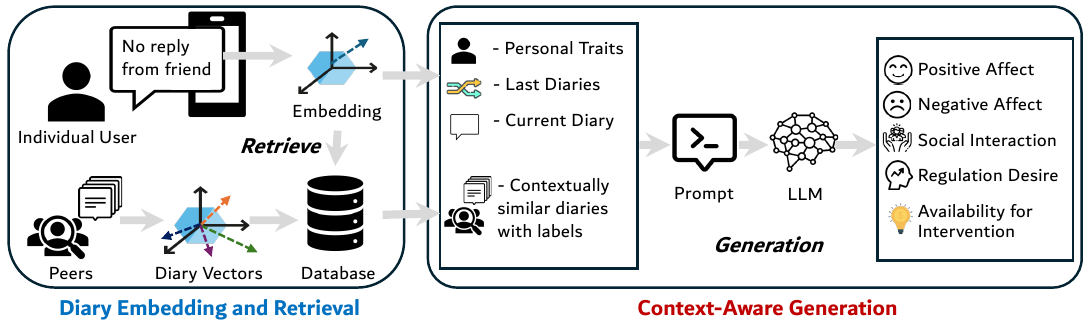}
    \caption{Illustration of the CALLM Framework: The left side represents the Diary Embedding and Retrieval Module, while the right side illustrates the Context-Aware Generation Module.}
    \label{fig:framework}
\end{figure*}

Building upon our contextual analysis findings from RQ1, we designed CALLM to leverage three key insights: (1) the significant impact of contextual factors on emotional states, (2) the substantial role of individual differences in emotional experiences, and (3) LLMs' pre-trained knowledge regarding cancer survivors' emotional experiences and potential needs. As illustrated in Figure~\ref{fig:framework}, the framework consists of two main modules: the Diary Embedding and Retrieval Module and the Context-Aware Generation Module. By embedding and retrieving contextually relevant emotion diary data from peers, along with integrating individual traits and historical trajectories, CALLM incorporates these contextual cues into the LLM generation process to identify current emotional and behavioral states. This integration improves both the interpretability and predictive accuracy for real-world emotional states, even from extremely brief diary entries.

\subsubsection{Diary Embedding and Retrieval Module}

The Diary Embedding and Retrieval Module begins by converting each diary entry into a high-dimensional text embedding using OpenAI's \texttt{text-embedding-3-small} model\footnote{\url{https://platform.openai.com/docs/guides/embeddings}{OpenAI - Vector Embeddings}}. These embeddings capture semantic similarities among entries and are stored in a vector database. Each vector is accompanied by metadata, including self-reported outcome scores and participant traits, ensuring rich contextual alignment.

To facilitate efficient retrieval, a FAISS (Facebook AI Similarity Search\footnote{\url{https://ai.meta.com/tools/faiss/}{Faiss: A library for efficient similarity search}}) index is constructed over these embeddings. This index allows for rapid similarity searches, leveraging L2 distance computation to identify the most relevant entries, where the L2 distance \(d(\mathbf{a}, \mathbf{b})\) between two vectors \(\mathbf{a}\) and \(\mathbf{b}\) is defined as: $
d(\mathbf{a}, \mathbf{b}) = \sqrt{\sum_{i=1}^{n} (a_i - b_i)^2}
$

When a new diary entry is embedded, the system queries the FAISS index to retrieve the top-$K$ semantically similar entries from the database. These entries are selected based on their proximity in the embedding space, indicating contextual relevance.

The retrieved entries, along with their associated ground truth labels, serve as contextual examples for the subsequent generation process. This retrieval mechanism ensures that predictions are grounded in contextually relevant historical data from peers, enhancing the model's ability to understand and predict emotional states in context despite entries containing limited content.

\subsubsection{Context-Aware Generation Module}

\begin{wrapfigure}{r}{0.4\linewidth}
    \centering
    \includegraphics[width=\linewidth]{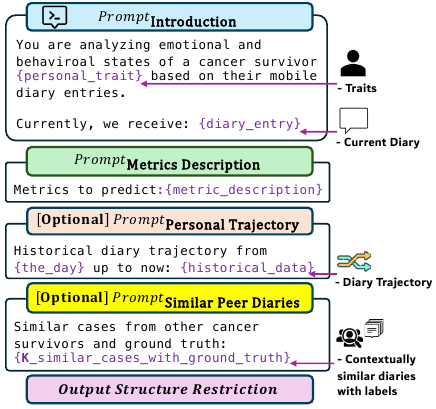}
    \caption{Illustration of simplified prompt design, with additional details in Figure \ref{fig:prompts} in Appendix \ref{apd:prompt}.}
    \label{fig:prompt_demo}
\end{wrapfigure}

The Context-Aware Generation Module integrates the retrieved data with additional contextual elements to improve prediction accuracy. As illustrated by Figure \ref{fig:prompt_demo}, this module constructs a comprehensive prompt by incorporating:

\begin{itemize}
    \item \textbf{Individual Traits:} Participant-specific demographic and clinical information (e.g., ``60-year-old male, stage II Kidney cancer'') to enable personalized analysis. These traits help contextualize the emotional experience within the individual's specific cancer journey and life circumstances, allowing for more nuanced interpretations of emotional states.
    \item \textbf{Temporal Trajectory Context:} Historical diary entries from the participant's own timeline (e.g., since ``Current Day'' or ``Last Day'', if available), constructing an emotional trajectory from their past diary entries. This temporal information helps capture emotional patterns and transitions, providing insight into how current emotions relate to recent experiences and potentially indicating emerging trends in emotional experiences.
    \item \textbf{Retrieved Peer Experiences:} The RAG module retrieves the k most similar cases (where k ranges from 1 to 20 in our experiments) using FAISS vector similarity search with L2 distance computation. Each retrieved case consists of an emotion diary text and its associated emotional outcomes, serving as concrete examples to ground the model's predictions.
\end{itemize}

This structured approach ensures that each prediction is grounded in both individual-specific context and broader patterns observed across the participant population. For each analysis, the new diary entry serves as the primary input, while the LLM is instructed to act as an emotion analysis assistant specifically trained for cancer survivors' emotional experiences. The prompt is formatted to generate predictions in a consistent JSON structure, enclosed within \texttt{<PREDICTIONS>} tags for reliable parsing. This standardized output format facilitates downstream processing and integration with other components of the emotion analysis pipeline.

\subsubsection{Inference and Postprocessing}
Using the constructed prompt, the LLM predicts both continuous affective scales and binary emotional states. The model outputs probabilities for each emotional state, ranging from 0.0 to 1.0, where 0.0 indicates the LLM assesses the specific state as unlikely, and 1.0 indicates it as likely. These probabilistic estimates are then converted into binary classifications using a 0.5 threshold. This step ensures compatibility with evaluation metrics and facilitates practical deployment of the framework. In our experiments, the framework utilizes \texttt{GPT-4o-mini} as the base LLM with 0.3 temperature and 1000 max tokens.

\subsection{Evaluation Methodology}
To assess whether CALLM can accurately predict emotional and behavioral states from brief mobile diary text (RQ2), we established a comprehensive evaluation protocol. We evaluated the framework using metrics including balanced accuracy (i.e., the mean of sensitivity and specificity) and area under the receiver operating characteristic curve (ROC-AUC) across binary emotional and behavioral states. Notably, both metrics maintain a naive baseline of guessing the majority as 50\% (referred as `majority baseline' later), reflecting the expected performance of majority guessing in binary classification, regardless of class imbalance. 

The evaluation was conducted using 5-fold grouped cross-validation, with each fold representing a distinct group of participants (which means participants were stratified into five non-overlapping groups, ensuring that all entries from the same participant remained within a single fold to prevent data leakage and maintain the independence of training and testing sets). This validation approach ensures that the model's performance metrics reflect its true generalizability to new, unseen participants rather than merely its ability to recognize patterns from familiar individuals.

We compared CALLM to several baseline approaches:
\begin{itemize}
\item[1)] \textbf{Traditional text classification methods:} We implemented Bag-of-Words \cite{harris1954distributional}, TF-IDF \cite{salton1988term}, and LIWC \cite{boyd2022development} feature extraction, each paired with logistic regression classifiers to enable simultaneous prediction across all target variables.
\item[2)] \textbf{General-purpose transformer models:} We fine-tuned state-of-the-art transformer architectures including BERT \cite{devlin2018bert}, SentenceBERT \cite{reimers2019sentence}, XLNet \cite{yang2019xlnet}, and RoBERTa \cite{liu2019roberta}. Each model was configured with shared representations and multiple classification heads to simultaneously predict all emotional/behavioral states, leveraging cross-task learning for improved performance.
\item[3)] \textbf{Emotion-specialized transformer models:} We evaluated pre-trained emotion-specific models including EmoBERT \cite{demszky2020goemotions} (trained on the GoEmotions dataset), RoBERTa-Emotion \cite{liu2019roberta}, DeBERTa-Emotion \cite{he2021deberta}, and Emotion-Transformer based on DistilBERT \cite{sanh2019distilbert}. These models were specifically pre-trained on emotion recognition tasks, providing a strong benchmark for affective text analysis.
\item[4)] \textbf{LLM-based classification:} We tested zero-shot and few-shot approaches using \texttt{`GPT-4o-mini'} (same as the base LLM for CALLM evaluation reported), with standard prompting and random few-shot retrieval from peers' diaries paired with ground-truth labels (versus semantic similarity-based retrieval used in CALLM). This allowed us to evaluate the effectiveness of CALLM's design of peer-experience-based retrieval augmentation and individual trajectory incorporation.
\end{itemize}

\begin{table*}[t]
    \small
    \centering
    \caption{Balanced Accuracy (\%, mean ± SD) on four prediction targets. 
Note, \texttt{LLM} rows show a vanilla \texttt{GPT-4o-mini} prompted with random 0/5/20-shot diaries paired with ground truth measures. 
\texttt{CALLM} rows add two enhancements: (i) similarity-based retrieval of $k$ peer diaries paired with ground truth measures (\emph{$k$-shot}) and (ii) optional personal diary history (\emph{Diary History} = None / Since Current Day / Since Last Day). 
Bold text indicates the best performance.}
    \label{tab:rag-shot-comparison}
    \begin{tabular}{ccccc}
    \toprule
    \textbf{Model} & 
    \textbf{PosAff (\%)} & 
    \textbf{NegAff (\%)} & 
    \textbf{RegDesire (\%)} & 
    \textbf{IntAvail (\%)} \\
    \midrule
    \texttt{Majority Baseline}      & 50.00$_{\scriptsize \pm 0.00}$ & 50.00$_{\scriptsize \pm 0.00}$ & 50.00$_{\scriptsize \pm 0.00}$ & 50.00$_{\scriptsize \pm 0.00}$ \\ 
    \midrule
    \texttt{BoW}          & 66.20$_{\scriptsize \pm 1.09}$ & 62.05$_{\scriptsize \pm 1.47}$ & 62.17$_{\scriptsize \pm 1.05}$ & 50.12$_{\scriptsize \pm 1.27}$ \\
    \texttt{TF-IDF}        & 67.34$_{\scriptsize \pm 1.67}$ & 62.75$_{\scriptsize \pm 1.16}$ & 62.52$_{\scriptsize \pm 0.83}$ & 50.69$_{\scriptsize \pm 1.29}$ \\
    \texttt{LIWC}        & 62.67$_{\scriptsize \pm 0.81}$ & 60.29$_{\scriptsize \pm 1.11}$ & 60.69$_{\scriptsize \pm 1.32}$ & 50.00$_{\scriptsize \pm 0.00}$ \\
    \texttt{XLNet}         & 61.55$_{\scriptsize \pm 0.91}$ & 57.98$_{\scriptsize \pm 1.14}$ & 59.75$_{\scriptsize \pm 1.38}$ & 52.20$_{\scriptsize \pm 0.82}$ \\
    \texttt{BERT}         & 67.99$_{\scriptsize \pm 1.38}$ & 64.39$_{\scriptsize \pm 2.15}$ & 64.81$_{\scriptsize \pm 1.29}$ & 51.42$_{\scriptsize \pm 1.14}$ \\
    \texttt{SentenceBERT} & 68.76$_{\scriptsize \pm 1.33}$ & 67.14$_{\scriptsize \pm 1.46}$ & 66.92$_{\scriptsize \pm 2.33}$ & 52.52$_{\scriptsize \pm 0.91}$ \\
    \texttt{EmoBERT}  & 65.81$_{\scriptsize \pm 1.18}$ & 64.18$_{\scriptsize \pm 1.38}$ & 65.54$_{\scriptsize \pm 2.03}$ & 52.41$_{\scriptsize \pm 0.92}$ \\
    \texttt{RoBERTa}  & 68.32$_{\scriptsize \pm 1.42}$ & 66.27$_{\scriptsize \pm 1.50}$ & 67.23$_{\scriptsize \pm 1.80}$ & 53.73$_{\scriptsize \pm 1.20}$ \\
    \texttt{RoBERTa-Emotion}  & 68.06$_{\scriptsize \pm 1.42}$ & 67.20$_{\scriptsize \pm 1.25}$ & 66.61$_{\scriptsize \pm 2.08}$ & 52.13$_{\scriptsize \pm 1.10}$ \\
    \texttt{DeBERTa-Emotion}  & 68.60$_{\scriptsize \pm 1.90}$ & 64.30$_{\scriptsize \pm 2.40}$ & 65.90$_{\scriptsize \pm 1.52}$ & 51.90$_{\scriptsize \pm 0.70}$ \\
    \texttt{Emotion-Transformer}  & 68.25$_{\scriptsize \pm 1.62}$ & 66.13$_{\scriptsize \pm 1.32}$ & 66.20$_{\scriptsize \pm 1.92}$ & 55.33$_{\scriptsize \pm 1.89}$ \\
    LLM Zero-shot  & 70.14$_{\scriptsize \pm 1.34}$ & 69.98$_{\scriptsize \pm 2.89}$ & 70.22$_{\scriptsize \pm 2.97}$ & 53.16$_{\scriptsize \pm 1.20}$  \\  
    LLM 5-Shot    & 69.95$_{\scriptsize \pm 1.42}$ & 69.87$_{\scriptsize \pm 2.76}$ & 70.10$_{\scriptsize \pm 3.05}$ & 53.05$_{\scriptsize \pm 1.18}$  \\
    LLM 20-Shot    & 69.99$_{\scriptsize \pm 1.50}$ & 69.75$_{\scriptsize \pm 2.90}$ & 69.95$_{\scriptsize \pm 2.95}$ & 52.98$_{\scriptsize \pm 1.22}$  \\
    \midrule
    \texttt{CALLM} 5-shot + None Diary       & 72.35$_{\scriptsize \pm 1.56}$ & 72.56$_{\scriptsize \pm 2.80}$ & 71.26$_{\scriptsize \pm 3.64}$ & 53.08$_{\scriptsize \pm 1.17}$ \\
    \texttt{CALLM} 0-shot + Diaries Since Current Day        & 70.25$_{\scriptsize \pm 1.37}$ & 72.18$_{\scriptsize \pm 2.69}$ & 73.15$_{\scriptsize \pm 3.22}$ & 55.55$_{\scriptsize \pm 1.22}$ \\
    \texttt{CALLM} 5-shot + Diaries Since Current Day      & 72.31$_{\scriptsize \pm 1.37}$ & 72.75$_{\scriptsize \pm 2.85}$ & 71.77$_{\scriptsize \pm 3.40}$ & 56.04$_{\scriptsize \pm 1.15}$ \\
    \texttt{CALLM} 0-shot + Diaries Since Last Day  & 70.42$_{\scriptsize \pm 1.47}$ & 72.30$_{\scriptsize \pm 2.68}$ & \textbf{73.72$_{\scriptsize \pm 3.40}$}  & 58.85$_{\scriptsize \pm 1.77}$ \\
    \texttt{CALLM} 5-shot + Diaries Since Last Day  & 72.26$_{\scriptsize \pm 1.46}$ & 72.65$_{\scriptsize \pm 2.65}$ & 71.45$_{\scriptsize \pm 3.60}$ & \textbf{60.09$_{\scriptsize \pm 1.15}$} \\
    \texttt{CALLM} 20-shot + Diaries Since Last Day & \textbf{72.96$_{\scriptsize \pm 1.54}$} & \textbf{73.29$_{\scriptsize \pm 2.89}$} & 71.86$_{\scriptsize \pm 3.12}$ & 56.31$_{\scriptsize \pm 1.23}$ \\
    \bottomrule
    \end{tabular}
\end{table*}

All models were evaluated under identical conditions using our 5-fold grouped cross-validation approach as the outer validation approach within nested cross validations for hyper-parameter tuning detailed in Appendix \ref{appendix:hyper-parameter}, ensuring a fair comparison across fundamentally different architectures and learning paradigms.

CALLM differs from the plain LLM baselines along two orthogonal axes:

\begin{enumerate}
    \item \textbf{Retrieval-augmented few-shot context:}  
    Rather than sampling examples at random, CALLM retrieves the $k\!\in\!\{0,5,20\}$ most semantically similar peer diaries paired with its ground-truth labels using FAISS similarity search.  Setting $k=0$ yields a pure zero-shot prompt, while $k>0$ conducts a similarity-based few-shot scheme, shown as `CALLM 0/5/20-shot' in Table \ref{tab:rag-shot-comparison}.
    \item \textbf{Personal temporal history:}  
    We optionally prepend the participant’s own recent diary trajectory so the model can exploit short-term temporal dependencies. We test three scopes: \emph{none}, \emph{entries since today’s morning}, and \emph{entries since yesterday’s morning}, shown as `None Diary', `Diaries Since Current Day (Morning)', and `Diary Since Last Day (Morning)' in Table \ref{tab:rag-shot-comparison}, respectively.
\end{enumerate}

Different combinations of these two factors yields the six CALLM variants in Table~\ref{tab:rag-shot-comparison}.  Crucially, every other setting—base model (\texttt{GPT-4o-mini}), temperature~$=0.3$, oken limit~$=1000$, and JSON output format—remains identical to the LLM baselines, allowing us to isolate the incremental benefit of
(i) similarity-based retrieval and (ii) personal history conditioning.

\subsection{Prediction Performance Results}

Table~\ref{tab:rag-shot-comparison} reports balanced--accuracy scores for all models on the four target constructs.  
Across the conventional baselines--including sparse text features paired with logistic regression and fine-tuned or emotion-specialised transformers--the highest performance plateaus at roughly $68$–$69\%$ for positive affect, $67$–$68\%$ for negative affect and regulation-desire, and $55\%$ for intervention availability.  
These figures are only a few points above the $50\%$ majority baseline of balanced accuracy, underscoring a clear ceiling when contextual information is absent.

For LLM baselines with no or random context awareness, the zero-shot setting lifts the upper bound for the three subjective constructs to $\approx70\%$, yet adding five or twenty randomly selected peer exemplars provides no additional benefit and in some folds yields a small decline. The lack of improvement suggests that exemplars unaligned with the index entry's situational context introduce distracting cues that offset any nominal few-shot advantage.

\begin{figure}[t!]
    \centering
    \begin{subfigure}{0.48\textwidth}
        \includegraphics[width=\textwidth]{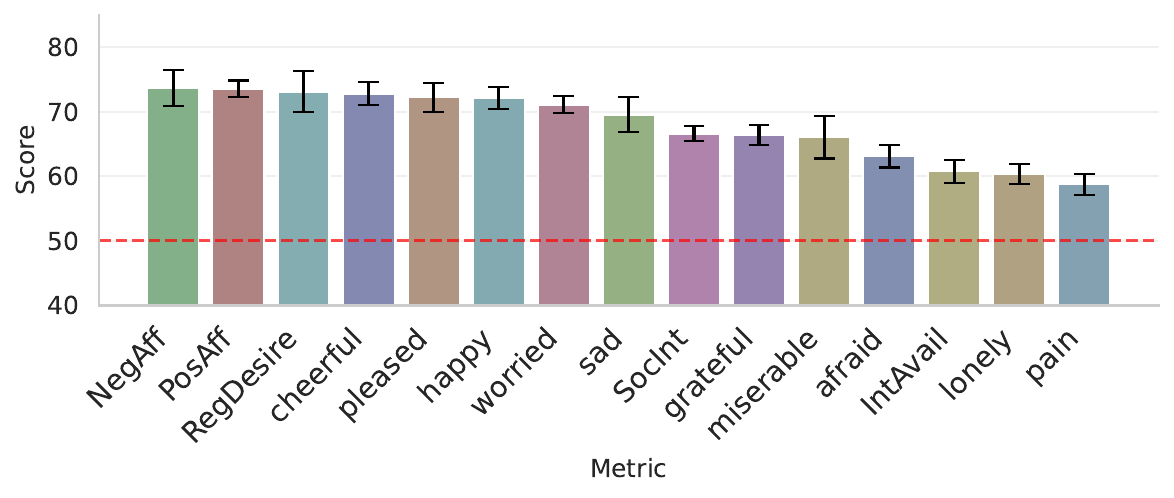}
        \caption{Balanced Accuracy.}
        \label{fig:balanced_accuracy}
    \end{subfigure}
    \hfill
    \begin{subfigure}{0.48\textwidth}
        \includegraphics[width=\textwidth]{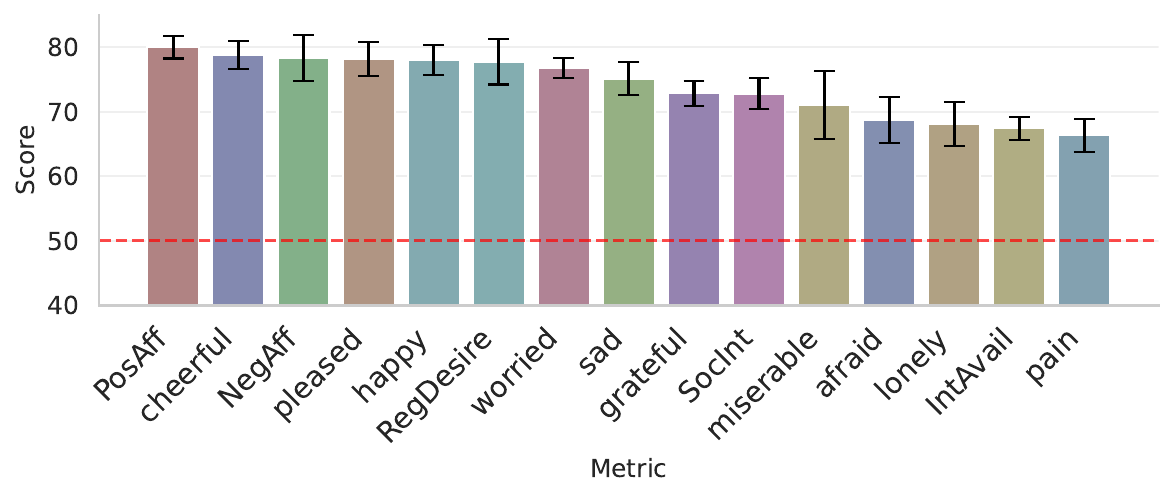}
        \caption{ROC-AUC.}
        \label{fig:roc_auc}
    \end{subfigure}
    
    \caption{Model performance across different emotional and behavioral states. The red dashed line indicates the naive majority baseline of 50\%.}
    \label{fig:model_performance_individual}
\end{figure}

In contrast, every \textsc{CALLM} variant surpasses baseline configurations.  
The best balanced-accuracy scores reach $72.96\%$ for positive affect, $73.29\%$ for negative affect, $73.72\%$ for regulation-desire, and $60.09\%$ for intervention availability, improving the ceiling compared to the baseline methods.
For affective states, the largest boost arises when similarity-matched peer exemplars are combined with the participant's diary trajectory since the prior day, implying that survivors share sufficiently convergent affective contexts for cross-participant retrieval to be informative.  
Regulation desire, however, improves significantly only when personal trajectory is present; adding peer examples alone offers no benefit and occasionally degrades performance, pointing to the idiosyncratic nature of moment-to-moment motivation to alter one's desire to regulate emotions.  
Availability detection remains the most challenging task, while \textsc{CALLM} raises accuracy to about $60\%$, indicating that this objective behavioral constraint may require additional multimodal signals beyond diary text.

Taken together, the results confirm that large language models can already decode brief diary entries more accurately than traditional text-classification pipelines, but that carefully curated contextual prompts—rather than randomly chosen exemplars—are essential for pushing performance beyond the $70~\%$ ceiling and for extracting the nuanced temporal cues embedded in participants’ recent histories.

\subsection{Generalizability Discussion}

To assess whether CALLM generalizes to the full spectrum of nuanced emotional and behavioral states captured in our dataset, we extended the evaluation to a wider set of constructs beyond the four core targets reported in Table~\ref{tab:rag-shot-comparison}.

As shown in Figure \ref{fig:model_performance_individual}, the model achieves balanced accuracies ranging from 58.8\% to 73.2\% and ROC-AUC scores from 66.4\% to 79.5\% across different states, consistently outperforming the majority baseline (50\%). Beyond the core constructs reported above, the model shows robust performance in detecting specific emotional states such as cheerfulness (72.8\%), pleasure (71.6\%), and worry (71.1\%). Even for more challenging states like fear (63.1\%), loneliness (60.4\%), and physical pain (58.8\%), the model maintains above-chance performance, highlighting its broad applicability across diverse emotional and behavioral dimensions in cancer survivors. 

\begin{figure*}[t]
    \centering
    \includegraphics[width=0.76\columnwidth]{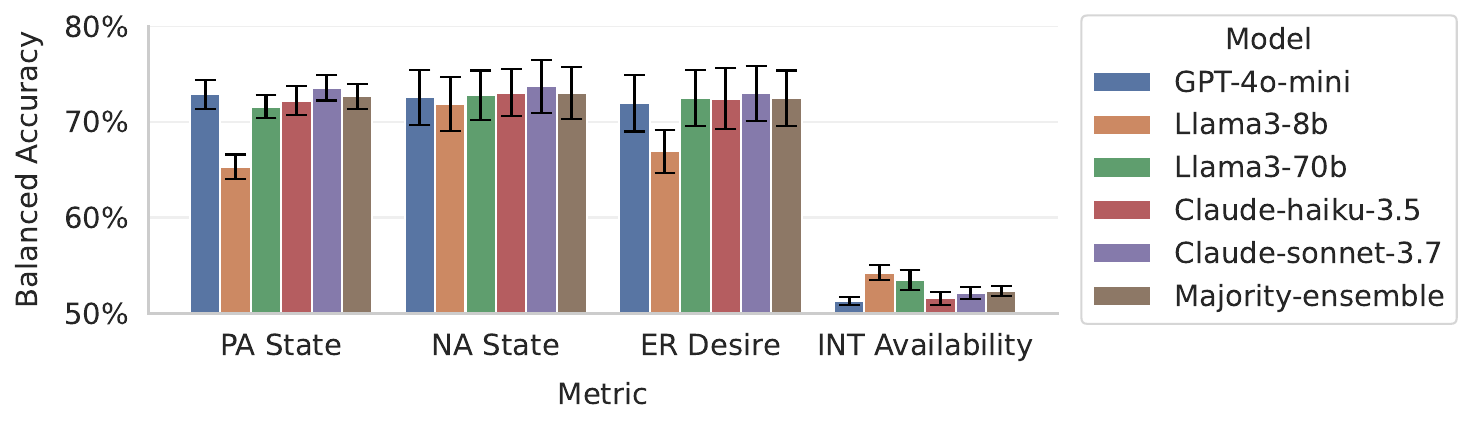}
    \caption{Model Performance Comparison across Metrics. Error bars represent standard deviation across 5-fold grouped cross-validation.}
    \label{fig:model_performance_comparison}
\end{figure*}

We also evaluated CALLM's generalizability across different base LLMs. See Figure \ref{fig:model_performance_comparison}, comparing GPT-4o-mini (OpenAI), Llama-3 models (8b and 70b), and Claude models (3.5-haiku and 3.7-sonnet), all models demonstrated consistent effectiveness in analyzing cancer survivors' diary entries. Ensembling the models (aggregating predictions from all models to conduct majority voting) does not contribute to better performance. Claude 3.7-sonnet slightly outperformed others across most constructs. Llama3-8b exhibited the lowest performance, perhaps due to the smaller model size which may not have enough power to capture the nuances of the data. Besides Llama3-8b, models' performance differences were minimal for most constructs (within 2-3\%), with high prediction result correlations between models (97\%), demonstrating CALLM's robustness across different LLM implementations.

\begin{wrapfigure}{tr}{0.4\linewidth}
    \centering
    \includegraphics[width=\linewidth]{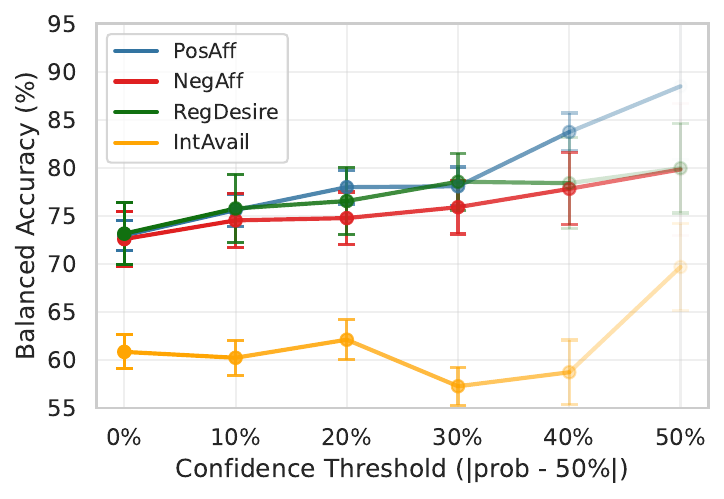}
    \caption{Balanced Accuracy performance for predicted states at different LLM confidence thresholds ($|{\rm probability} - 50\%|$; ranging from 0 to 50\%, higher confidence thresholds yield better accuracy but fewer samples). Transparency indicates sample retention rate; error bars show standard deviations across 5-fold grouped cross-validation.}
    \label{fig:confidence_threshold_analysis}
\end{wrapfigure}

\section{Analysis of Performance Determinants}

To address \textbf{RQ3} concerning the factors that influence model performance, we conducted several post-hoc analyses exploring how model confidence, diary entry length, temporal context, and personalization affect prediction accuracy. These analyses provide valuable insights for real-world deployment of LLM-based mobile diary analysis systems.

\begin{wrapfigure}{t}{0.7\linewidth}
    \centering
    \includegraphics[width=\linewidth]{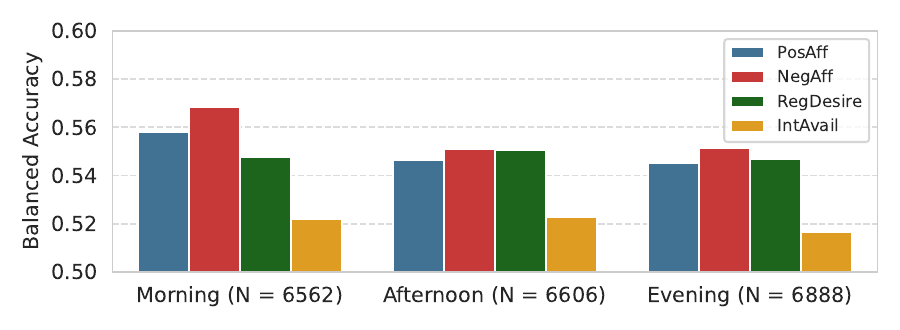}
    \caption{Next-day prediction performance by time period. The figure shows balanced accuracy across different emotional and behavioral states when predicting morning, afternoon, and evening states of the following day.}
    \label{fig:next-day-prediction}
\end{wrapfigure}

\subsection{Model Confidence and Prediction Accuracy}
Our first analysis examined the relationship between the LLM's confidence (probability estimates) and prediction accuracy. By varying confidence thresholds (defined as $|{\rm probability} - 50\%|$), we found that all states except intervention availability showed improved prediction accuracy as confidence increased. For positive affect, accuracy improved dramatically from 73\% to 88\% at the highest confidence levels (Figure~\ref{fig:confidence_threshold_analysis}). This finding suggests that in practical applications, the system could use confidence scores to selectively trigger interventions only when predictions reach a certain reliability threshold.

\subsection{Effect of Diary Entry Length}
We also investigated how diary entry length affects prediction performance by analyzing entries containing 3-15 words. Spearman rank correlations revealed that for most emotional states (9 out of 15 analyzed), longer entries significantly improved prediction accuracy. Positive affect showed a strong positive correlation ($\rho = 0.874$, adjusted $p < 0.01$) with entry length, while intervention availability demonstrated a strong negative correlation ($\rho = -0.996$, adjusted $p < 0.001$), suggesting that shorter entries may be more effective for this particular prediction. Negative affect showed no significant relationship with text length ($\rho = -0.276$, adjusted $p = 0.472$). These findings provide practical guidance for optimizing prompt length in mobile diary applications, potentially with construct-specific instructions.

\subsection{Temporal Prediction Capabilities}
To explore the potential for forecasting emotional states—valuable for proactive intervention—we tested whether current-day diary entries could predict next-day emotional states. The analysis showed modest but consistent predictive power, with balanced accuracies between 54-56\% across constructs, significantly above chance level (Wilcoxon signed-rank test, adjusted $p < 0.001$). As shown in Figure~\ref{fig:next-day-prediction}, prediction accuracy varied by time of day, with morning states being marginally more predictable than afternoon or evening periods. This temporal pattern suggests that emotional states may be more stable and predictable in the morning, possibly before the accumulated effects of daily stressors. While the predictive capability is limited, it offers potential for anticipatory intervention planning, particularly for morning-focused interventions based on the previous day's diary content.

\subsection{Benefits of Personalization}
Finally, we examined whether incorporating individual-level ground truth data could enhance prediction accuracy. By simulating a scenario where the system collects one week of user self-reports before operating independently, we found statistically significant improvements for key emotional states: positive affect (1.53\% improvement, adjusted $p < 0.05$), negative affect (1.42\% improvement, adjusted $p < 0.05$), and social interaction quality (3.62\% improvement, adjusted $p < 0.01$). These results suggest that a short calibration period can meaningfully improve the personalization of emotional state predictions in a practical deployment scenario.

\section{Discussion}

Here we discuss the findings and takeaways from the present study with respect to the research questions, clinical implications we inferred from the findings, practical deployment considerations towards actionable intervention systems.

\subsection{Findings and takeaways}

Our research addressed three key questions about analyzing cancer survivors' mobile diary entries. For \textbf{RQ1}, our analysis revealed systematic relationships between described contexts and emotional states—administrative and health-related contexts were associated with greater negative affect and regulation needs, while leisure and social activities promoted positive affect. Even without explicit emotional language, these brief diary entries reflected cancer survivors' unique experiences and indicated their emotional states. Importantly, intervention availability varied significantly across contexts (higher at home, lower during work/study), highlighting opportunities for context-sensitive support. The significant proportion of variance explained by individual random effects underscores the importance of personalization, as participants often responded differently under similar contextual circumstances.

For \textbf{RQ2}, we demonstrated that CALLM successfully leverages both peer experiences and temporal diary context to analyze emotional states from brief diary entries, enabling mobile health applications to predict these states from the emotion drivers found in diary entries alone. This approach minimizes user burden by harnessing the information in one brief diary response rather than having users repeatedly report on their current emotional and behavioral states across many distinct questionnaire items. By combining LLMs' pre-trained knowledge, personal diary history, with RAG-based peer exemplars, CALLM delivers flexible, context-aware predictions--even from sparse data—consistently outperformed baselines with balanced accuracies of 72.96\% for positive affect, 73.29\% for negative affect, and 73.72\% for emotion regulation desire. Its consistent performance across various emotional constructs attests to both its effectiveness and generalizability. Moreover, the RAG-enabled peers' contextually-similar exemplar retrieval is particularly valuable in privacy-sensitive healthcare settings, as it enables learning from an anonymized database of peer experiences and personal diary history maintained on local devices.

Addressing \textbf{RQ3}, our post-hoc analyses revealed four practical insights that influence prediction performance: (1) Model confidence strongly predicts accuracy, with improvements up to 15 percentage points when filtering for high-confidence predictions, suggesting confidence thresholds in deployed systems could significantly enhance reliability. (2) Longer diary entries generally enhance predictive performance for affective states, but overly long entries offer diminishing returns, indicating a trade-off between information gain and user burden—especially when analytical efficiency is prioritized. This observation aligns with prior findings on optimizing assessment length \cite{10.1145/3699735}.
(3) Diaries provide modest yet consistent predictive value for next-day emotional states, particularly for morning periods, suggesting value for short-term rather than long-range forecasting. and (4) Brief personalization periods—such as collecting one week of self-reported ground truth measures and incorporating them into the retrieval process—can yield meaningful improvements in prediction accuracy, particularly for emotional states.

Both our pre-hoc and post-hoc analyses demonstrate that context-aware language models can illuminate the complex, dynamic emotional experiences of cancer survivors through their open-ended diary entries. These findings collectively demonstrate the potential of extracting rich contextual understanding from brief mobile diary entries to inform timing, content, and personalization of mobile interventions while respecting individual differences in emotional responses to similar contexts—with important implications for developing more effective, less burdensome mobile health applications.

\subsection{Clinical Implications}

DMHIs can increase access to needed mental health care for cancer survivors by overcoming the significant barriers to care that they otherwise face (e.g., social stigma \cite{holland2010psychosocial}, financial costs \cite{niazi2020barriers}, time burdens \cite{bradbury2019developing}, and lack of available providers \cite{camm2019people}). Despite cancer survivors finding digital interventions acceptable \cite{corbett2018understanding}, sustained engagement with digital interventions is a widespread problem \cite{fleming2018beyond}. Factors that can impact user engagement relate to the timing of interventions, their dosage, and the burden placed on individuals to use them. Just-in-time adaptive intervention (JITAI) frameworks offer a way to improve intervention engagement and effectiveness by increasing the personal relevance of intervention timing, content, and dosage \cite{nahum2018just}. However, researchers have yet to identify a convenient and low-burden method for identifying when to intervene and what intervention to provide to patients. This is in part due to a tension between gathering enough information from participants to accurately inform when to provide an intervention, and a logical desire to minimize burden by not requiring them to report on their emotions multiple times per day using the same questionnaire items. Minimizing user burden is critical to mitigate the risk of individuals losing interest and dropping out of an intervention altogether.

The CALLM framework proposed here represents an important step towards reducing user burden without sacrificing the personalized, contextual information required to guide optimized intervention delivery decisions. For instance, we demonstrate that brief emotion-driver diary entries—each comprising just a few words—can predict whether an individual is feeling more negative than usual, more positive than usual, or desires to change their emotional state with approximately 73\% accuracy, rising further when LLMs report higher confidence levels. Although digital diaries have been widely adopted for emotional self-regulation, symptom monitoring, and reflective journaling, CALLM extends these capabilities by providing a lightweight channel for real-time inference of user states and intervention opportunities. Moreover, passive sensing of behavioral and contextual cues—such as GPS-derived location (e.g., workspace, home, or outdoor), ambient sound patterns (indicating social activity versus solitude), and phone-usage logs (reflecting availability for interaction)—offers an objective complement to subjective desire measurements \cite{harari2023understanding}. By integrating these signals, CALLM can potentially supplement the roughly 60\% accuracy limitation of intervention-availability detection based solely on diaries, enabling more precisely timed, personalized interventions that enhance both relevance and engagement.

The ability to reasonably approximate these emotion factors and intervention opportunities in survivors through their brief diaries is a drastic improvement over the traditional method of asking individuals about each of these states in separate questionnaire items, to determine whether an intervention is warranted. Furthermore, our findings suggest complementary integration with passive sensing data (such as location, activity, and device usage patterns) could create even more comprehensive context awareness, as we found systematic relationships between certain contexts (e.g., home activities, administrative tasks) and intervention opportunity components. Such multi-modal approaches could further reduce user burden while improving prediction accuracy, especially for constructs like intervention availability that showed more modest prediction performance from text alone. CALLM's potential to improve JITAI delivery systems while minimizing user input is an exciting path forward to better address the mental health needs of cancer survivors.

\subsection{Deployment Considerations}

Healthcare applications of emotion analysis require privacy safeguards under HIPAA \cite{act1996health} and GDPR \cite{regulation2018general}, including encrypted storage, clear consent processes, and data minimization practices. Our RAG-based approach offers an additional privacy advantage by using anonymized peer experiences rather than requiring extensive personal data collection, aligning with privacy-by-design principles.

Computational efficiency is another practical consideration. Taking the GPT-4o-mini as an example, processing costs (\$0.0005 per diary entry) remain feasible at scale, with potential for further reduction through smaller, specialized models. Recent advances with sub-billion parameter models like MobileLLM \cite{liu2024mobilellm} suggest on-device deployment is viable, enhancing both privacy and accessibility.

Implementing our post-hoc insights would further optimize real-world deployment. For instance, adaptive confidence thresholds could dynamically balance prediction reliability against coverage, triggering interventions only when confidence exceeds certain thresholds. Similarly, contextual guidance about optimal diary entry length could improve user experience while maintaining prediction quality. For healthcare systems integration, the framework could provide confidence scores alongside predictions, giving clinicians transparency into reliability when reviewing automated assessments. These practical considerations help bridge the gap between research findings and sustainable implementation in clinical settings.

\subsection{Future Work}

This study represents a first step in using mobile health data and LLMs to identify cancer survivors' emotional and behavioral states and inform personalized DMHIs, with limitations to acknowledge which open up future directions. It is worth noting that our study utilized GPT-4o-mini as the base LLM - while this demonstrates the potential of even smaller language models in healthcare applications, a comprehensive benchmark of different LLM architectures was not our focus and remains an open direction for future research.

Notably, future work should address demographic limitations of our study, with data skewed toward female (90.17\%) and White (86.67\%) participants. Extending validation to more diverse populations is critical, as recent research suggests language models may perform differently across demographic groups \cite{blodgett2020language}, potentially affecting equitable emotion analysis. 

While this study focuses on proactive diary entries, future work could incorporate complementary ``passive sensing'' data such as phone usage, location, and physiological signals \cite{harari2023understanding,wang2023detecting}. Such passive data could supplement analysis when diary entries are unavailable, creating a more resilient system while minimizing user burden. Additionally, while the diary entries were collected in participants’ everyday lives, participants were compensated for completing the study. As such, future diary collection efforts without participant compensation may encounter different engagement patterns.

Additionally, while LLMs demonstrate impressive performance in our emotion analysis tasks, their ``black box'' nature presents challenges for clinical applications where understanding prediction rationale is essential. Future work should explore better interpretability through prediction explanations, user-oriented feedback loop, and alignment with established clinical scales. Meanwhile, our finding that model confidence correlates strongly with prediction accuracy offers a practical mechanism for communicating reliability to clinicians and patients.

Towards more personalized emotion management solutions, such as JITAIs \cite{wu2024mindshift} and embodied conversational chatbot \cite{chiu2024computational}, the next step might be longer-term context studies to understand how LLM systems adapt to changing emotional patterns and life circumstances of cancer survivors. Future work should explore mechanisms for continuous learning and adaptation of the model to individual users over extended periods, particularly in handling evolving emotional needs and identifying shifting patterns in intervention opportunities during different phases of survivorship. We envision this study as a feasibility test toward building LLM-empowered agentic workflows that incorporate memory of individual users' context trajectories, sensor- and diary-based context understanding, retrieval-augmented external knowledge integration, and theory-informed workflow automation, leveraging LLM systems' neural-symbolic and task execution potential \cite{xiong2024converging,stade2024large}. This could ultimately lead to more personalized and context-aware emotional management solutions for cancer survivors and potentially broader populations.

\section{Conclusion}

This paper introduced CALLM, a context-aware framework for analyzing brief mobile diary entries from cancer survivors. Using only diary entries collected via smartphones and validated with data from 407 cancer survivors, CALLM performed well in predicting emotional states and variables relevant to personalized intervention opportunities (desire to regulate emotions and availability to engage in an interventions). By uniquely integrating retrieved peer experiences with individual temporal trajectories, our framework demonstrates how contextual information in mobile diaries can be effectively leveraged to understand users' emotional experiences (RQ1), predict key states from brief text with reasonable accuracy (RQ2), and identify factors that optimize prediction performance like confidence thresholds and diary length (RQ3). These contributions highlight how context-aware computational approaches can transform lightweight user inputs into rich behavioral insights while addressing the critical challenge of identifying optimal moments for intervention delivery—potentially reducing assessment burden while enabling more responsive and personalized mobile health applications.

\bibliographystyle{ACM-Reference-Format}
\bibliography{reference}


\begin{thebibliography}{95}


\ifx \showCODEN    \undefined \def \showCODEN     #1{\unskip}     \fi
\ifx \showISBNx    \undefined \def \showISBNx     #1{\unskip}     \fi
\ifx \showISBNxiii \undefined \def \showISBNxiii  #1{\unskip}     \fi
\ifx \showISSN     \undefined \def \showISSN      #1{\unskip}     \fi
\ifx \showLCCN     \undefined \def \showLCCN      #1{\unskip}     \fi
\ifx \shownote     \undefined \def \shownote      #1{#1}          \fi
\ifx \showarticletitle \undefined \def \showarticletitle #1{#1}   \fi
\ifx \showURL      \undefined \def \showURL       {\relax}        \fi
\providecommand\bibfield[2]{#2}
\providecommand\bibinfo[2]{#2}
\providecommand\natexlab[1]{#1}
\providecommand\showeprint[2][]{arXiv:#2}

\bibitem[Abowd et~al\mbox{.}(1999)]%
        {abowd1999towards}
\bibfield{author}{\bibinfo{person}{Gregory~D Abowd}, \bibinfo{person}{Anind~K
  Dey}, \bibinfo{person}{Peter~J Brown}, \bibinfo{person}{Nigel Davies},
  \bibinfo{person}{Mark Smith}, {and} \bibinfo{person}{Pete Steggles}.}
  \bibinfo{year}{1999}\natexlab{}.
\newblock \showarticletitle{Towards a better understanding of context and
  context-awareness}. In \bibinfo{booktitle}{\emph{Handheld and Ubiquitous
  Computing: First International Symposium, HUC’99 Karlsruhe, Germany,
  September 27--29, 1999 Proceedings 1}}. Springer, \bibinfo{pages}{304--307}.
\newblock


\bibitem[Act(1996)]%
        {act1996health}
\bibfield{author}{\bibinfo{person}{Accountability Act}.}
  \bibinfo{year}{1996}\natexlab{}.
\newblock \showarticletitle{Health insurance portability and accountability act
  of 1996}.
\newblock \bibinfo{journal}{\emph{Public law}}  \bibinfo{volume}{104}
  (\bibinfo{year}{1996}), \bibinfo{pages}{191}.
\newblock


\bibitem[Alshurafa et~al\mbox{.}(2018)]%
        {10.1145/3287031}
\bibfield{author}{\bibinfo{person}{Nabil Alshurafa},
  \bibinfo{person}{Jayalakshmi Jain}, \bibinfo{person}{Rawan Alharbi},
  \bibinfo{person}{Gleb Iakovlev}, \bibinfo{person}{Bonnie Spring}, {and}
  \bibinfo{person}{Angela Pfammatter}.} \bibinfo{year}{2018}\natexlab{}.
\newblock \showarticletitle{Is More Always Better? Discovering Incentivized
  mHealth Intervention Engagement Related to Health Behavior Trends}.
\newblock \bibinfo{journal}{\emph{Proc. ACM Interact. Mob. Wearable Ubiquitous
  Technol.}} \bibinfo{volume}{2}, \bibinfo{number}{4}, Article
  \bibinfo{articleno}{153} (\bibinfo{date}{Dec.} \bibinfo{year}{2018}),
  \bibinfo{numpages}{26}~pages.
\newblock
\href{https://doi.org/10.1145/3287031}{doi:\nolinkurl{10.1145/3287031}}


\bibitem[Andrykowski et~al\mbox{.}(2008)]%
        {andrykowski2008psychological}
\bibfield{author}{\bibinfo{person}{Michael~A Andrykowski},
  \bibinfo{person}{Emily Lykins}, {and} \bibinfo{person}{Andrea Floyd}.}
  \bibinfo{year}{2008}\natexlab{}.
\newblock \showarticletitle{Psychological health in cancer survivors}. In
  \bibinfo{booktitle}{\emph{Seminars in oncology nursing}},
  Vol.~\bibinfo{volume}{24}. Elsevier, \bibinfo{pages}{193--201}.
\newblock


\bibitem[Blodgett et~al\mbox{.}(2020)]%
        {blodgett2020language}
\bibfield{author}{\bibinfo{person}{Su~Lin Blodgett}, \bibinfo{person}{Solon
  Barocas}, \bibinfo{person}{Hal Daum{\'e}~III}, {and} \bibinfo{person}{Hanna
  Wallach}.} \bibinfo{year}{2020}\natexlab{}.
\newblock \showarticletitle{Language (technology) is power: A critical survey
  of" bias" in nlp}.
\newblock \bibinfo{journal}{\emph{arXiv preprint arXiv:2005.14050}}
  (\bibinfo{year}{2020}).
\newblock


\bibitem[Bolger et~al\mbox{.}(2003)]%
        {bolger2003diary}
\bibfield{author}{\bibinfo{person}{Niall Bolger}, \bibinfo{person}{Angelina
  Davis}, {and} \bibinfo{person}{Eshkol Rafaeli}.}
  \bibinfo{year}{2003}\natexlab{}.
\newblock \showarticletitle{Diary methods: Capturing life as it is lived}.
\newblock \bibinfo{journal}{\emph{Annual review of psychology}}
  \bibinfo{volume}{54}, \bibinfo{number}{1} (\bibinfo{year}{2003}),
  \bibinfo{pages}{579--616}.
\newblock


\bibitem[Boyd et~al\mbox{.}(2022)]%
        {boyd2022development}
\bibfield{author}{\bibinfo{person}{Ryan~L Boyd}, \bibinfo{person}{Ashwini
  Ashokkumar}, \bibinfo{person}{Sarah Seraj}, {and} \bibinfo{person}{James~W
  Pennebaker}.} \bibinfo{year}{2022}\natexlab{}.
\newblock \showarticletitle{The development and psychometric properties of
  LIWC-22}.
\newblock \bibinfo{journal}{\emph{Austin, TX: University of Texas at Austin}}
  \bibinfo{volume}{10} (\bibinfo{year}{2022}).
\newblock


\bibitem[Bradbury et~al\mbox{.}(2019)]%
        {bradbury2019developing}
\bibfield{author}{\bibinfo{person}{Katherine Bradbury}, \bibinfo{person}{Mary
  Steele}, \bibinfo{person}{Teresa Corbett}, \bibinfo{person}{Adam~WA
  Geraghty}, \bibinfo{person}{Adele Krusche}, \bibinfo{person}{Elena Heber},
  \bibinfo{person}{Steph Easton}, \bibinfo{person}{Tara Cheetham-Blake},
  \bibinfo{person}{Joanna Slodkowska-Barabasz}, \bibinfo{person}{Andre~Matthias
  M{\"u}ller}, {et~al\mbox{.}}} \bibinfo{year}{2019}\natexlab{}.
\newblock \showarticletitle{Developing a digital intervention for cancer
  survivors: an evidence-, theory-and person-based approach}.
\newblock \bibinfo{journal}{\emph{NPJ digital medicine}} \bibinfo{volume}{2},
  \bibinfo{number}{1} (\bibinfo{year}{2019}), \bibinfo{pages}{85}.
\newblock


\bibitem[Businelle et~al\mbox{.}(2020)]%
        {businelle2020reducing}
\bibfield{author}{\bibinfo{person}{Michael~S Businelle},
  \bibinfo{person}{Scott~T Walters}, \bibinfo{person}{Eun-Young Mun},
  \bibinfo{person}{Thomas~R Kirchner}, \bibinfo{person}{Emily~T H{\'e}bert},
  {and} \bibinfo{person}{Xiaoyin Li}.} \bibinfo{year}{2020}\natexlab{}.
\newblock \showarticletitle{Reducing drinking among people experiencing
  homelessness: protocol for the development and testing of a just-in-time
  adaptive intervention}.
\newblock \bibinfo{journal}{\emph{JMIR Research Protocols}}
  \bibinfo{volume}{9}, \bibinfo{number}{4} (\bibinfo{year}{2020}),
  \bibinfo{pages}{e15610}.
\newblock


\bibitem[Camm-Crosbie et~al\mbox{.}(2019)]%
        {camm2019people}
\bibfield{author}{\bibinfo{person}{Louise Camm-Crosbie},
  \bibinfo{person}{Louise Bradley}, \bibinfo{person}{Rebecca Shaw},
  \bibinfo{person}{Simon Baron-Cohen}, {and} \bibinfo{person}{Sarah Cassidy}.}
  \bibinfo{year}{2019}\natexlab{}.
\newblock \showarticletitle{‘People like me don’t get support’: Autistic
  adults’ experiences of support and treatment for mental health
  difficulties, self-injury and suicidality}.
\newblock \bibinfo{journal}{\emph{Autism}} \bibinfo{volume}{23},
  \bibinfo{number}{6} (\bibinfo{year}{2019}), \bibinfo{pages}{1431--1441}.
\newblock


\bibitem[Chen et~al\mbox{.}(2023)]%
        {chen2023empowering}
\bibfield{author}{\bibinfo{person}{Zhiyu Chen}, \bibinfo{person}{Yujie Lu},
  {and} \bibinfo{person}{William~Yang Wang}.} \bibinfo{year}{2023}\natexlab{}.
\newblock \showarticletitle{Empowering psychotherapy with large language
  models: Cognitive distortion detection through diagnosis of thought
  prompting}.
\newblock \bibinfo{journal}{\emph{arXiv preprint arXiv:2310.07146}}
  (\bibinfo{year}{2023}).
\newblock


\bibitem[Chiu et~al\mbox{.}(2024)]%
        {chiu2024computational}
\bibfield{author}{\bibinfo{person}{Yu~Ying Chiu}, \bibinfo{person}{Ashish
  Sharma}, \bibinfo{person}{Inna~Wanyin Lin}, {and} \bibinfo{person}{Tim
  Althoff}.} \bibinfo{year}{2024}\natexlab{}.
\newblock \showarticletitle{A computational framework for behavioral assessment
  of llm therapists}.
\newblock \bibinfo{journal}{\emph{arXiv preprint arXiv:2401.00820}}
  (\bibinfo{year}{2024}).
\newblock


\bibitem[Clifford et~al\mbox{.}(2018)]%
        {clifford2018barriers}
\bibfield{author}{\bibinfo{person}{Briana~K Clifford}, \bibinfo{person}{David
  Mizrahi}, \bibinfo{person}{Carolina~X Sandler}, \bibinfo{person}{Benjamin~K
  Barry}, \bibinfo{person}{David Simar}, \bibinfo{person}{Claire~E Wakefield},
  {and} \bibinfo{person}{David Goldstein}.} \bibinfo{year}{2018}\natexlab{}.
\newblock \showarticletitle{Barriers and facilitators of exercise experienced
  by cancer survivors: a mixed methods systematic review}.
\newblock \bibinfo{journal}{\emph{Supportive care in cancer}}
  \bibinfo{volume}{26} (\bibinfo{year}{2018}), \bibinfo{pages}{685--700}.
\newblock


\bibitem[Corbett et~al\mbox{.}(2018)]%
        {corbett2018understanding}
\bibfield{author}{\bibinfo{person}{Teresa Corbett}, \bibinfo{person}{Karmpaul
  Singh}, \bibinfo{person}{Liz Payne}, \bibinfo{person}{Katherine Bradbury},
  \bibinfo{person}{Claire Foster}, \bibinfo{person}{Eila Watson},
  \bibinfo{person}{Alison Richardson}, \bibinfo{person}{Paul Little}, {and}
  \bibinfo{person}{Lucy Yardley}.} \bibinfo{year}{2018}\natexlab{}.
\newblock \showarticletitle{Understanding acceptability of and engagement with
  Web-based interventions aiming to improve quality of life in cancer
  survivors: a synthesis of current research}.
\newblock \bibinfo{journal}{\emph{Psycho-oncology}} \bibinfo{volume}{27},
  \bibinfo{number}{1} (\bibinfo{year}{2018}), \bibinfo{pages}{22--33}.
\newblock


\bibitem[Davies et~al\mbox{.}(2023)]%
        {davies2023individual}
\bibfield{author}{\bibinfo{person}{Andy Davies}, \bibinfo{person}{Eiko Fried},
  \bibinfo{person}{Omar Costilla-Reyes}, {and} \bibinfo{person}{Hane Aung}.}
  \bibinfo{year}{2023}\natexlab{}.
\newblock \showarticletitle{Individual Behavioral Insights in Schizophrenia: A
  Network Analysis and Mobile Sensing Approach}. In
  \bibinfo{booktitle}{\emph{International Conference on Pervasive Computing
  Technologies for Healthcare}}. Springer, \bibinfo{pages}{18--33}.
\newblock


\bibitem[De~Vries et~al\mbox{.}(2021)]%
        {de2021smartphone}
\bibfield{author}{\bibinfo{person}{Lianne~P De~Vries}, \bibinfo{person}{Bart~ML
  Baselmans}, {and} \bibinfo{person}{Meike Bartels}.}
  \bibinfo{year}{2021}\natexlab{}.
\newblock \showarticletitle{Smartphone-based ecological momentary assessment of
  well-being: A systematic review and recommendations for future studies}.
\newblock \bibinfo{journal}{\emph{Journal of Happiness Studies}}
  \bibinfo{volume}{22}, \bibinfo{number}{5} (\bibinfo{year}{2021}),
  \bibinfo{pages}{2361--2408}.
\newblock


\bibitem[Demszky et~al\mbox{.}(2020)]%
        {demszky2020goemotions}
\bibfield{author}{\bibinfo{person}{Dorottya Demszky}, \bibinfo{person}{Dana
  Movshovitz-Attias}, \bibinfo{person}{Jeongwoo Ko}, \bibinfo{person}{Alan
  Cowen}, \bibinfo{person}{Gaurav Nemade}, {and} \bibinfo{person}{Sujith
  Ravi}.} \bibinfo{year}{2020}\natexlab{}.
\newblock \showarticletitle{GoEmotions: A Dataset of Fine-Grained Emotions}. In
  \bibinfo{booktitle}{\emph{Proceedings of the 58th Annual Meeting of the
  Association for Computational Linguistics}}. \bibinfo{pages}{4040--4054}.
\newblock


\bibitem[Devlin(2018)]%
        {devlin2018bert}
\bibfield{author}{\bibinfo{person}{Jacob Devlin}.}
  \bibinfo{year}{2018}\natexlab{}.
\newblock \showarticletitle{Bert: Pre-training of deep bidirectional
  transformers for language understanding}.
\newblock \bibinfo{journal}{\emph{arXiv preprint arXiv:1810.04805}}
  (\bibinfo{year}{2018}).
\newblock


\bibitem[Eichstaedt et~al\mbox{.}(2018)]%
        {eichstaedt2018facebook}
\bibfield{author}{\bibinfo{person}{Johannes~C Eichstaedt},
  \bibinfo{person}{Robert~J Smith}, \bibinfo{person}{Raina~M Merchant},
  \bibinfo{person}{Lyle~H Ungar}, \bibinfo{person}{Patrick Crutchley},
  \bibinfo{person}{Daniel Preo{\c{t}}iuc-Pietro}, \bibinfo{person}{David~A
  Asch}, {and} \bibinfo{person}{H~Andrew Schwartz}.}
  \bibinfo{year}{2018}\natexlab{}.
\newblock \showarticletitle{Facebook language predicts depression in medical
  records}.
\newblock \bibinfo{journal}{\emph{Proceedings of the National Academy of
  Sciences}} \bibinfo{volume}{115}, \bibinfo{number}{44}
  (\bibinfo{year}{2018}), \bibinfo{pages}{11203--11208}.
\newblock


\bibitem[Firkins et~al\mbox{.}(2020)]%
        {firkins2020quality}
\bibfield{author}{\bibinfo{person}{Jenny Firkins}, \bibinfo{person}{Lissi
  Hansen}, \bibinfo{person}{Martha Driessnack}, {and} \bibinfo{person}{Nathan
  Dieckmann}.} \bibinfo{year}{2020}\natexlab{}.
\newblock \showarticletitle{Quality of life in “chronic” cancer survivors:
  a meta-analysis}.
\newblock \bibinfo{journal}{\emph{Journal of Cancer Survivorship}}
  \bibinfo{volume}{14} (\bibinfo{year}{2020}), \bibinfo{pages}{504--517}.
\newblock


\bibitem[Fleming et~al\mbox{.}(2018)]%
        {fleming2018beyond}
\bibfield{author}{\bibinfo{person}{Theresa Fleming}, \bibinfo{person}{Lynda
  Bavin}, \bibinfo{person}{Mathijs Lucassen}, \bibinfo{person}{Karolina
  Stasiak}, \bibinfo{person}{Sarah Hopkins}, {and} \bibinfo{person}{Sally
  Merry}.} \bibinfo{year}{2018}\natexlab{}.
\newblock \showarticletitle{Beyond the trial: systematic review of real-world
  uptake and engagement with digital self-help interventions for depression,
  low mood, or anxiety}.
\newblock \bibinfo{journal}{\emph{Journal of medical Internet research}}
  \bibinfo{volume}{20}, \bibinfo{number}{6} (\bibinfo{year}{2018}),
  \bibinfo{pages}{e199}.
\newblock


\bibitem[Funk et~al\mbox{.}(2020)]%
        {funk2020framework}
\bibfield{author}{\bibinfo{person}{Burkhardt Funk}, \bibinfo{person}{Shiri
  Sadeh-Sharvit}, \bibinfo{person}{Ellen~E Fitzsimmons-Craft},
  \bibinfo{person}{Mickey~Todd Trockel}, \bibinfo{person}{Grace~E Monterubio},
  \bibinfo{person}{Neha~J Goel}, \bibinfo{person}{Katherine~N Balantekin},
  \bibinfo{person}{Dawn~M Eichen}, \bibinfo{person}{Rachael~E Flatt},
  \bibinfo{person}{Marie-Laure Firebaugh}, {et~al\mbox{.}}}
  \bibinfo{year}{2020}\natexlab{}.
\newblock \showarticletitle{A framework for applying natural language
  processing in digital health interventions}.
\newblock \bibinfo{journal}{\emph{Journal of medical Internet research}}
  \bibinfo{volume}{22}, \bibinfo{number}{2} (\bibinfo{year}{2020}),
  \bibinfo{pages}{e13855}.
\newblock


\bibitem[Gao et~al\mbox{.}(2023)]%
        {10.1145/3580798}
\bibfield{author}{\bibinfo{person}{Chenyang Gao}, \bibinfo{person}{Ivan
  Marsic}, \bibinfo{person}{Aleksandra Sarcevic}, \bibinfo{person}{Waverly
  Gestrich-Thompson}, {and} \bibinfo{person}{Randall~S. Burd}.}
  \bibinfo{year}{2023}\natexlab{}.
\newblock \showarticletitle{Real-time Context-Aware Multimodal Network for
  Activity and Activity-Stage Recognition from Team Communication in Dynamic
  Clinical Settings}.
\newblock \bibinfo{journal}{\emph{Proc. ACM Interact. Mob. Wearable Ubiquitous
  Technol.}} \bibinfo{volume}{7}, \bibinfo{number}{1}, Article
  \bibinfo{articleno}{12} (\bibinfo{date}{March} \bibinfo{year}{2023}),
  \bibinfo{numpages}{28}~pages.
\newblock
\href{https://doi.org/10.1145/3580798}{doi:\nolinkurl{10.1145/3580798}}


\bibitem[Guntuku et~al\mbox{.}(2019)]%
        {guntuku2019understanding}
\bibfield{author}{\bibinfo{person}{Sharath~Chandra Guntuku},
  \bibinfo{person}{Anneke Buffone}, \bibinfo{person}{Kokil Jaidka},
  \bibinfo{person}{Johannes~C Eichstaedt}, {and} \bibinfo{person}{Lyle~H
  Ungar}.} \bibinfo{year}{2019}\natexlab{}.
\newblock \showarticletitle{Understanding and measuring psychological stress
  using social media}. In \bibinfo{booktitle}{\emph{Proceedings of the
  international AAAI conference on web and social media}},
  Vol.~\bibinfo{volume}{13}. \bibinfo{pages}{214--225}.
\newblock


\bibitem[Harari and Gosling(2023)]%
        {harari2023understanding}
\bibfield{author}{\bibinfo{person}{Gabriella~M Harari} {and}
  \bibinfo{person}{Samuel~D Gosling}.} \bibinfo{year}{2023}\natexlab{}.
\newblock \showarticletitle{Understanding behaviours in context using mobile
  sensing}.
\newblock \bibinfo{journal}{\emph{Nature Reviews Psychology}}
  \bibinfo{volume}{2}, \bibinfo{number}{12} (\bibinfo{year}{2023}),
  \bibinfo{pages}{767--779}.
\newblock


\bibitem[Hardeman et~al\mbox{.}(2019)]%
        {hardeman2019systematic}
\bibfield{author}{\bibinfo{person}{Wendy Hardeman}, \bibinfo{person}{Julie
  Houghton}, \bibinfo{person}{Kathleen Lane}, \bibinfo{person}{Andy Jones},
  {and} \bibinfo{person}{Felix Naughton}.} \bibinfo{year}{2019}\natexlab{}.
\newblock \showarticletitle{A systematic review of just-in-time adaptive
  interventions (JITAIs) to promote physical activity}.
\newblock \bibinfo{journal}{\emph{International Journal of Behavioral Nutrition
  and Physical Activity}}  \bibinfo{volume}{16} (\bibinfo{year}{2019}),
  \bibinfo{pages}{1--21}.
\newblock


\bibitem[Harris(1954)]%
        {harris1954distributional}
\bibfield{author}{\bibinfo{person}{ZS Harris}.}
  \bibinfo{year}{1954}\natexlab{}.
\newblock \bibinfo{title}{Distributional structure}.
\newblock


\bibitem[He et~al\mbox{.}(2021)]%
        {he2021deberta}
\bibfield{author}{\bibinfo{person}{Pengcheng He}, \bibinfo{person}{Xiaodong
  Liu}, \bibinfo{person}{Jianfeng Gao}, {and} \bibinfo{person}{Weizhu Chen}.}
  \bibinfo{year}{2021}\natexlab{}.
\newblock \showarticletitle{DEBERTA: DECODING-ENHANCED BERT WITH DISENTANGLED
  ATTENTION}. In \bibinfo{booktitle}{\emph{International Conference on Learning
  Representations}}.
\newblock
\urldef\tempurl%
\url{https://openreview.net/forum?id=XPZIaotutsD}
\showURL{%
\tempurl}


\bibitem[Hewitt and Rowland(2002)]%
        {hewitt2002mental}
\bibfield{author}{\bibinfo{person}{Maria Hewitt} {and} \bibinfo{person}{Julia~H
  Rowland}.} \bibinfo{year}{2002}\natexlab{}.
\newblock \showarticletitle{Mental health service use among adult cancer
  survivors: analyses of the National Health Interview Survey}.
\newblock \bibinfo{journal}{\emph{Journal of Clinical Oncology}}
  \bibinfo{volume}{20}, \bibinfo{number}{23} (\bibinfo{year}{2002}),
  \bibinfo{pages}{4581--4590}.
\newblock


\bibitem[Holland et~al\mbox{.}(2010)]%
        {holland2010psychosocial}
\bibfield{author}{\bibinfo{person}{Jimmie~C Holland}, \bibinfo{person}{Brian~J
  Kelly}, {and} \bibinfo{person}{Mark~I Weinberger}.}
  \bibinfo{year}{2010}\natexlab{}.
\newblock \showarticletitle{Why psychosocial care is difficult to integrate
  into routine cancer care: stigma is the elephant in the room}.
\newblock \bibinfo{journal}{\emph{Journal of the National Comprehensive Cancer
  Network}} \bibinfo{volume}{8}, \bibinfo{number}{4} (\bibinfo{year}{2010}),
  \bibinfo{pages}{362--366}.
\newblock


\bibitem[Jacobs and Shulman(2017)]%
        {jacobs2017follow}
\bibfield{author}{\bibinfo{person}{Linda~A Jacobs} {and}
  \bibinfo{person}{Lawrence~N Shulman}.} \bibinfo{year}{2017}\natexlab{}.
\newblock \showarticletitle{Follow-up care of cancer survivors: challenges and
  solutions}.
\newblock \bibinfo{journal}{\emph{The Lancet Oncology}} \bibinfo{volume}{18},
  \bibinfo{number}{1} (\bibinfo{year}{2017}), \bibinfo{pages}{e19--e29}.
\newblock


\bibitem[Jung et~al\mbox{.}(2024)]%
        {10.1145/3675094.3677601}
\bibfield{author}{\bibinfo{person}{Gyeyoung Jung}, \bibinfo{person}{Soyeon
  Choi}, \bibinfo{person}{Yuju Kang}, {and} \bibinfo{person}{Jaejeung Kim}.}
  \bibinfo{year}{2024}\natexlab{}.
\newblock \showarticletitle{MyListener: An AI-Mediated Journaling Mobile
  Application for Alleviating Depression and Loneliness Using Contextual Data}.
  In \bibinfo{booktitle}{\emph{Companion of the 2024 on ACM International Joint
  Conference on Pervasive and Ubiquitous Computing}} (Melbourne VIC, Australia)
  \emph{(\bibinfo{series}{UbiComp '24})}. \bibinfo{publisher}{Association for
  Computing Machinery}, \bibinfo{address}{New York, NY, USA},
  \bibinfo{pages}{137–141}.
\newblock
\showISBNx{9798400710582}
\href{https://doi.org/10.1145/3675094.3677601}{doi:\nolinkurl{10.1145/3675094.3677601}}


\bibitem[Kim et~al\mbox{.}(2020)]%
        {10.1145/3432213}
\bibfield{author}{\bibinfo{person}{Inyeop Kim}, \bibinfo{person}{Hwarang Goh},
  \bibinfo{person}{Nematjon Narziev}, \bibinfo{person}{Youngtae Noh}, {and}
  \bibinfo{person}{Uichin Lee}.} \bibinfo{year}{2020}\natexlab{}.
\newblock \showarticletitle{Understanding User Contexts and Coping Strategies
  for Context-aware Phone Distraction Management System Design}.
\newblock \bibinfo{journal}{\emph{Proc. ACM Interact. Mob. Wearable Ubiquitous
  Technol.}} \bibinfo{volume}{4}, \bibinfo{number}{4}, Article
  \bibinfo{articleno}{134} (\bibinfo{date}{Dec.} \bibinfo{year}{2020}),
  \bibinfo{numpages}{33}~pages.
\newblock
\href{https://doi.org/10.1145/3432213}{doi:\nolinkurl{10.1145/3432213}}


\bibitem[Kim et~al\mbox{.}(2024)]%
        {10.1145/3613904.3642937}
\bibfield{author}{\bibinfo{person}{Taewan Kim}, \bibinfo{person}{Seolyeong
  Bae}, \bibinfo{person}{Hyun~Ah Kim}, \bibinfo{person}{Su-Woo Lee},
  \bibinfo{person}{Hwajung Hong}, \bibinfo{person}{Chanmo Yang}, {and}
  \bibinfo{person}{Young-Ho Kim}.} \bibinfo{year}{2024}\natexlab{}.
\newblock \showarticletitle{MindfulDiary: Harnessing Large Language Model to
  Support Psychiatric Patients' Journaling}. In
  \bibinfo{booktitle}{\emph{Proceedings of the 2024 CHI Conference on Human
  Factors in Computing Systems}} (Honolulu, HI, USA)
  \emph{(\bibinfo{series}{CHI '24})}. \bibinfo{publisher}{Association for
  Computing Machinery}, \bibinfo{address}{New York, NY, USA}, Article
  \bibinfo{articleno}{701}, \bibinfo{numpages}{20}~pages.
\newblock
\showISBNx{9798400703300}
\href{https://doi.org/10.1145/3613904.3642937}{doi:\nolinkurl{10.1145/3613904.3642937}}


\bibitem[Klasnja and Pratt(2012)]%
        {klasnja2012healthcare}
\bibfield{author}{\bibinfo{person}{Predrag Klasnja} {and}
  \bibinfo{person}{Wanda Pratt}.} \bibinfo{year}{2012}\natexlab{}.
\newblock \showarticletitle{Healthcare in the pocket: mapping the space of
  mobile-phone health interventions}.
\newblock \bibinfo{journal}{\emph{Journal of biomedical informatics}}
  \bibinfo{volume}{45}, \bibinfo{number}{1} (\bibinfo{year}{2012}),
  \bibinfo{pages}{184--198}.
\newblock


\bibitem[Kleiman et~al\mbox{.}(2018)]%
        {kleiman2018digital}
\bibfield{author}{\bibinfo{person}{Evan~M Kleiman}, \bibinfo{person}{Brianna~J
  Turner}, \bibinfo{person}{Szymon Fedor}, \bibinfo{person}{Eleanor~E Beale},
  \bibinfo{person}{Rosalind~W Picard}, \bibinfo{person}{Jeff~C Huffman}, {and}
  \bibinfo{person}{Matthew~K Nock}.} \bibinfo{year}{2018}\natexlab{}.
\newblock \showarticletitle{Digital phenotyping of suicidal thoughts}.
\newblock \bibinfo{journal}{\emph{Depression and anxiety}}
  \bibinfo{volume}{35}, \bibinfo{number}{7} (\bibinfo{year}{2018}),
  \bibinfo{pages}{601--608}.
\newblock


\bibitem[Krieke et~al\mbox{.}(2016)]%
        {krieke2016hownutsarethedutch}
\bibfield{author}{\bibinfo{person}{Lian Van~Der Krieke},
  \bibinfo{person}{Bertus~F Jeronimus}, \bibinfo{person}{Frank~J Blaauw},
  \bibinfo{person}{Rob~BK Wanders}, \bibinfo{person}{Ando~C Emerencia},
  \bibinfo{person}{Hendrika~M Schenk}, \bibinfo{person}{Stijn~De Vos},
  \bibinfo{person}{Evelien Snippe}, \bibinfo{person}{Marieke Wichers},
  \bibinfo{person}{Johanna~TW Wigman}, {et~al\mbox{.}}}
  \bibinfo{year}{2016}\natexlab{}.
\newblock \showarticletitle{HowNutsAreTheDutch (HoeGekIsNL): A crowdsourcing
  study of mental symptoms and strengths}.
\newblock \bibinfo{journal}{\emph{International journal of methods in
  psychiatric research}} \bibinfo{volume}{25}, \bibinfo{number}{2}
  (\bibinfo{year}{2016}), \bibinfo{pages}{123--144}.
\newblock


\bibitem[Lee et~al\mbox{.}(2024)]%
        {10.1145/3699723}
\bibfield{author}{\bibinfo{person}{Jong~Ho Lee}, \bibinfo{person}{Sunghoon~Ivan
  Lee}, {and} \bibinfo{person}{Eun~Kyoung Choe}.}
  \bibinfo{year}{2024}\natexlab{}.
\newblock \showarticletitle{GoalTrack: Supporting Personalized Goal-Setting in
  Stroke Rehabilitation with Multimodal Activity Journaling}.
\newblock \bibinfo{journal}{\emph{Proc. ACM Interact. Mob. Wearable Ubiquitous
  Technol.}} \bibinfo{volume}{8}, \bibinfo{number}{4}, Article
  \bibinfo{articleno}{167} (\bibinfo{date}{Nov.} \bibinfo{year}{2024}),
  \bibinfo{numpages}{29}~pages.
\newblock
\href{https://doi.org/10.1145/3699723}{doi:\nolinkurl{10.1145/3699723}}


\bibitem[Lewis et~al\mbox{.}(2020)]%
        {lewis2020retrieval}
\bibfield{author}{\bibinfo{person}{Patrick Lewis}, \bibinfo{person}{Ethan
  Perez}, \bibinfo{person}{Aleksandra Piktus}, \bibinfo{person}{Fabio Petroni},
  \bibinfo{person}{Vladimir Karpukhin}, \bibinfo{person}{Naman Goyal},
  \bibinfo{person}{Heinrich K{\"u}ttler}, \bibinfo{person}{Mike Lewis},
  \bibinfo{person}{Wen-tau Yih}, \bibinfo{person}{Tim Rockt{\"a}schel},
  {et~al\mbox{.}}} \bibinfo{year}{2020}\natexlab{}.
\newblock \showarticletitle{Retrieval-augmented generation for
  knowledge-intensive nlp tasks}.
\newblock \bibinfo{journal}{\emph{Advances in Neural Information Processing
  Systems}}  \bibinfo{volume}{33} (\bibinfo{year}{2020}),
  \bibinfo{pages}{9459--9474}.
\newblock


\bibitem[Li et~al\mbox{.}(2024)]%
        {10.1145/3699735}
\bibfield{author}{\bibinfo{person}{Jixin Li}, \bibinfo{person}{Aditya Ponnada},
  \bibinfo{person}{Wei-Lin Wang}, \bibinfo{person}{Genevieve Dunton}, {and}
  \bibinfo{person}{Stephen Intille}.} \bibinfo{year}{2024}\natexlab{}.
\newblock \showarticletitle{Ask Less, Learn More: Adapting Ecological Momentary
  Assessment Survey Length by Modeling Question-Answer Information Gain}.
\newblock \bibinfo{journal}{\emph{Proc. ACM Interact. Mob. Wearable Ubiquitous
  Technol.}} \bibinfo{volume}{8}, \bibinfo{number}{4}, Article
  \bibinfo{articleno}{166} (\bibinfo{date}{Nov.} \bibinfo{year}{2024}),
  \bibinfo{numpages}{32}~pages.
\newblock
\href{https://doi.org/10.1145/3699735}{doi:\nolinkurl{10.1145/3699735}}


\bibitem[Linton et~al\mbox{.}(2021)]%
        {linton2021investigating}
\bibfield{author}{\bibinfo{person}{Myles-Jay~Anthony Linton},
  \bibinfo{person}{Sarah Jelbert}, \bibinfo{person}{Judi Kidger},
  \bibinfo{person}{Richard Morris}, \bibinfo{person}{Lucy Biddle}, {and}
  \bibinfo{person}{Bruce Hood}.} \bibinfo{year}{2021}\natexlab{}.
\newblock \showarticletitle{Investigating the use of electronic well-being
  diaries completed within a psychoeducation program for university students:
  Longitudinal text analysis study}.
\newblock \bibinfo{journal}{\emph{Journal of medical Internet research}}
  \bibinfo{volume}{23}, \bibinfo{number}{4} (\bibinfo{year}{2021}),
  \bibinfo{pages}{e25279}.
\newblock


\bibitem[Liu et~al\mbox{.}(2019)]%
        {liu2019roberta}
\bibfield{author}{\bibinfo{person}{Yinhan Liu}, \bibinfo{person}{Myle Ott},
  \bibinfo{person}{Naman Goyal}, \bibinfo{person}{Jingfei Du},
  \bibinfo{person}{Mandar Joshi}, \bibinfo{person}{Danqi Chen},
  \bibinfo{person}{Omer Levy}, \bibinfo{person}{Mike Lewis},
  \bibinfo{person}{Luke Zettlemoyer}, {and} \bibinfo{person}{Veselin
  Stoyanov}.} \bibinfo{year}{2019}\natexlab{}.
\newblock \showarticletitle{Roberta: A robustly optimized bert pretraining
  approach}.
\newblock \bibinfo{journal}{\emph{arXiv preprint arXiv:1907.11692}}
  (\bibinfo{year}{2019}).
\newblock


\bibitem[Liu et~al\mbox{.}(2024)]%
        {liu2024mobilellm}
\bibfield{author}{\bibinfo{person}{Zechun Liu}, \bibinfo{person}{Changsheng
  Zhao}, \bibinfo{person}{Forrest Iandola}, \bibinfo{person}{Chen Lai},
  \bibinfo{person}{Yuandong Tian}, \bibinfo{person}{Igor Fedorov},
  \bibinfo{person}{Yunyang Xiong}, \bibinfo{person}{Ernie Chang},
  \bibinfo{person}{Yangyang Shi}, \bibinfo{person}{Raghuraman Krishnamoorthi},
  {et~al\mbox{.}}} \bibinfo{year}{2024}\natexlab{}.
\newblock \showarticletitle{Mobilellm: Optimizing sub-billion parameter
  language models for on-device use cases}. In
  \bibinfo{booktitle}{\emph{Forty-first International Conference on Machine
  Learning}}.
\newblock


\bibitem[Loria et~al\mbox{.}(2018)]%
        {loria2018textblob}
\bibfield{author}{\bibinfo{person}{Steven Loria} {et~al\mbox{.}}}
  \bibinfo{year}{2018}\natexlab{}.
\newblock \showarticletitle{textblob Documentation}.
\newblock \bibinfo{journal}{\emph{Release 0.15}} \bibinfo{volume}{2},
  \bibinfo{number}{8} (\bibinfo{year}{2018}), \bibinfo{pages}{269}.
\newblock


\bibitem[Ma et~al\mbox{.}(2024)]%
        {ma2024understanding}
\bibfield{author}{\bibinfo{person}{Zilin Ma}, \bibinfo{person}{Yiyang Mei},
  {and} \bibinfo{person}{Zhaoyuan Su}.} \bibinfo{year}{2024}\natexlab{}.
\newblock \showarticletitle{Understanding the benefits and challenges of using
  large language model-based conversational agents for mental well-being
  support}. In \bibinfo{booktitle}{\emph{AMIA Annual Symposium Proceedings}},
  Vol.~\bibinfo{volume}{2023}. \bibinfo{pages}{1105}.
\newblock


\bibitem[Mayer et~al\mbox{.}(2017)]%
        {mayer2017defining}
\bibfield{author}{\bibinfo{person}{Deborah~K Mayer},
  \bibinfo{person}{Shelly~Fuld Nasso}, {and} \bibinfo{person}{Jo~Anne Earp}.}
  \bibinfo{year}{2017}\natexlab{}.
\newblock \showarticletitle{Defining cancer survivors, their needs, and
  perspectives on survivorship health care in the USA}.
\newblock \bibinfo{journal}{\emph{The Lancet Oncology}} \bibinfo{volume}{18},
  \bibinfo{number}{1} (\bibinfo{year}{2017}), \bibinfo{pages}{e11--e18}.
\newblock


\bibitem[Mets{\"a}ranta et~al\mbox{.}(2019)]%
        {metsaranta2019adolescents}
\bibfield{author}{\bibinfo{person}{Kiki Mets{\"a}ranta}, \bibinfo{person}{Marjo
  Kurki}, \bibinfo{person}{Maritta Valimaki}, {and} \bibinfo{person}{Minna
  Anttila}.} \bibinfo{year}{2019}\natexlab{}.
\newblock \showarticletitle{How do adolescents use electronic diaries? A
  mixed-methods study among adolescents with depressive symptoms}.
\newblock \bibinfo{journal}{\emph{Journal of Medical Internet Research}}
  \bibinfo{volume}{21}, \bibinfo{number}{2} (\bibinfo{year}{2019}),
  \bibinfo{pages}{e11711}.
\newblock


\bibitem[Mohr et~al\mbox{.}(2017)]%
        {mohr2017personal}
\bibfield{author}{\bibinfo{person}{David~C Mohr}, \bibinfo{person}{Mi Zhang},
  {and} \bibinfo{person}{Stephen~M Schueller}.}
  \bibinfo{year}{2017}\natexlab{}.
\newblock \showarticletitle{Personal sensing: understanding mental health using
  ubiquitous sensors and machine learning}.
\newblock \bibinfo{journal}{\emph{Annual review of clinical psychology}}
  \bibinfo{volume}{13}, \bibinfo{number}{1} (\bibinfo{year}{2017}),
  \bibinfo{pages}{23--47}.
\newblock


\bibitem[Morris et~al\mbox{.}(2010)]%
        {morris2010mobile}
\bibfield{author}{\bibinfo{person}{Margaret~E Morris}, \bibinfo{person}{Qusai
  Kathawala}, \bibinfo{person}{Todd~K Leen}, \bibinfo{person}{Ethan~E
  Gorenstein}, \bibinfo{person}{Farzin Guilak}, \bibinfo{person}{Michael
  Labhard}, {and} \bibinfo{person}{William Deleeuw}.}
  \bibinfo{year}{2010}\natexlab{}.
\newblock \showarticletitle{Mobile therapy: case study evaluations of a cell
  phone application for emotional self-awareness}.
\newblock \bibinfo{journal}{\emph{Journal of medical Internet research}}
  \bibinfo{volume}{12}, \bibinfo{number}{2} (\bibinfo{year}{2010}),
  \bibinfo{pages}{e10}.
\newblock


\bibitem[Mustafa et~al\mbox{.}(2022)]%
        {mustafa2022user}
\bibfield{author}{\bibinfo{person}{Abdulsalam~Salihu Mustafa},
  \bibinfo{person}{Nor’ashikin Ali}, \bibinfo{person}{Jaspaljeet~Singh
  Dhillon}, \bibinfo{person}{Gamal Alkawsi}, {and} \bibinfo{person}{Yahia
  Baashar}.} \bibinfo{year}{2022}\natexlab{}.
\newblock \showarticletitle{User engagement and abandonment of mHealth: a
  cross-sectional survey}. In \bibinfo{booktitle}{\emph{Healthcare}},
  Vol.~\bibinfo{volume}{10}. MDPI, \bibinfo{pages}{221}.
\newblock


\bibitem[Nahum-Shani et~al\mbox{.}(2018)]%
        {nahum2018just}
\bibfield{author}{\bibinfo{person}{Inbal Nahum-Shani},
  \bibinfo{person}{Shawna~N Smith}, \bibinfo{person}{Bonnie~J Spring},
  \bibinfo{person}{Linda~M Collins}, \bibinfo{person}{Katie Witkiewitz},
  \bibinfo{person}{Ambuj Tewari}, {and} \bibinfo{person}{Susan~A Murphy}.}
  \bibinfo{year}{2018}\natexlab{}.
\newblock \showarticletitle{Just-in-time adaptive interventions (JITAIs) in
  mobile health: key components and design principles for ongoing health
  behavior support}.
\newblock \bibinfo{journal}{\emph{Annals of Behavioral Medicine}}
  (\bibinfo{year}{2018}), \bibinfo{pages}{1--17}.
\newblock


\bibitem[Nepal et~al\mbox{.}(2024a)]%
        {10.1145/3613905.3650767}
\bibfield{author}{\bibinfo{person}{Subigya Nepal}, \bibinfo{person}{Arvind
  Pillai}, \bibinfo{person}{William Campbell}, \bibinfo{person}{Talie
  Massachi}, \bibinfo{person}{Eunsol~Soul Choi}, \bibinfo{person}{Xuhai Xu},
  \bibinfo{person}{Joanna Kuc}, \bibinfo{person}{Jeremy~F Huckins},
  \bibinfo{person}{Jason Holden}, \bibinfo{person}{Colin Depp},
  \bibinfo{person}{Nicholas Jacobson}, \bibinfo{person}{Mary~P Czerwinski},
  \bibinfo{person}{Eric Granholm}, {and} \bibinfo{person}{Andrew Campbell}.}
  \bibinfo{year}{2024}\natexlab{a}.
\newblock \showarticletitle{Contextual AI Journaling: Integrating LLM and Time
  Series Behavioral Sensing Technology to Promote Self-Reflection and
  Well-being using the MindScape App}. In \bibinfo{booktitle}{\emph{Extended
  Abstracts of the CHI Conference on Human Factors in Computing Systems}}
  (Honolulu, HI, USA) \emph{(\bibinfo{series}{CHI EA '24})}.
  \bibinfo{publisher}{Association for Computing Machinery},
  \bibinfo{address}{New York, NY, USA}, Article \bibinfo{articleno}{86},
  \bibinfo{numpages}{8}~pages.
\newblock
\showISBNx{9798400703317}
\href{https://doi.org/10.1145/3613905.3650767}{doi:\nolinkurl{10.1145/3613905.3650767}}


\bibitem[Nepal et~al\mbox{.}(2024b)]%
        {10.1145/3699761}
\bibfield{author}{\bibinfo{person}{Subigya Nepal}, \bibinfo{person}{Arvind
  Pillai}, \bibinfo{person}{William Campbell}, \bibinfo{person}{Talie
  Massachi}, \bibinfo{person}{Michael~V. Heinz}, \bibinfo{person}{Ashmita
  Kunwar}, \bibinfo{person}{Eunsol~Soul Choi}, \bibinfo{person}{Xuhai Xu},
  \bibinfo{person}{Joanna Kuc}, \bibinfo{person}{Jeremy~F. Huckins},
  \bibinfo{person}{Jason Holden}, \bibinfo{person}{Sarah~M. Preum},
  \bibinfo{person}{Colin Depp}, \bibinfo{person}{Nicholas Jacobson},
  \bibinfo{person}{Mary~P. Czerwinski}, \bibinfo{person}{Eric Granholm}, {and}
  \bibinfo{person}{Andrew~T. Campbell}.} \bibinfo{year}{2024}\natexlab{b}.
\newblock \showarticletitle{MindScape Study: Integrating LLM and Behavioral
  Sensing for Personalized AI-Driven Journaling Experiences}.
\newblock \bibinfo{journal}{\emph{Proc. ACM Interact. Mob. Wearable Ubiquitous
  Technol.}} \bibinfo{volume}{8}, \bibinfo{number}{4}, Article
  \bibinfo{articleno}{186} (\bibinfo{date}{Nov.} \bibinfo{year}{2024}),
  \bibinfo{numpages}{44}~pages.
\newblock
\href{https://doi.org/10.1145/3699761}{doi:\nolinkurl{10.1145/3699761}}


\bibitem[Niazi et~al\mbox{.}(2020)]%
        {niazi2020barriers}
\bibfield{author}{\bibinfo{person}{Shehzad Niazi}, \bibinfo{person}{Emily
  Vargas}, \bibinfo{person}{Aaron Spaulding}, \bibinfo{person}{Elaine
  Gustetic}, \bibinfo{person}{Nancy Ford}, \bibinfo{person}{David Paly},
  \bibinfo{person}{Kelsey Tatum}, \bibinfo{person}{Matthew~M Clark}, {and}
  \bibinfo{person}{Teresa Rummans}.} \bibinfo{year}{2020}\natexlab{}.
\newblock \showarticletitle{Barriers to accepting mental health care in cancer
  patients with depression}.
\newblock \bibinfo{journal}{\emph{Social Work in Health Care}}
  \bibinfo{volume}{59}, \bibinfo{number}{6} (\bibinfo{year}{2020}),
  \bibinfo{pages}{351--364}.
\newblock


\bibitem[Nobles et~al\mbox{.}(2018)]%
        {10.1145/3173574.3173987}
\bibfield{author}{\bibinfo{person}{Alicia~L. Nobles},
  \bibinfo{person}{Jeffrey~J. Glenn}, \bibinfo{person}{Kamran Kowsari},
  \bibinfo{person}{Bethany~A. Teachman}, {and} \bibinfo{person}{Laura~E.
  Barnes}.} \bibinfo{year}{2018}\natexlab{}.
\newblock \showarticletitle{Identification of Imminent Suicide Risk Among Young
  Adults using Text Messages}. In \bibinfo{booktitle}{\emph{Proceedings of the
  2018 CHI Conference on Human Factors in Computing Systems}} (Montreal QC,
  Canada) \emph{(\bibinfo{series}{CHI '18})}. \bibinfo{publisher}{Association
  for Computing Machinery}, \bibinfo{address}{New York, NY, USA},
  \bibinfo{pages}{1–11}.
\newblock
\showISBNx{9781450356206}
\href{https://doi.org/10.1145/3173574.3173987}{doi:\nolinkurl{10.1145/3173574.3173987}}


\bibitem[Papacharissi(2010)]%
        {papacharissi2010networked}
\bibfield{author}{\bibinfo{person}{Zizi Papacharissi}.}
  \bibinfo{year}{2010}\natexlab{}.
\newblock \bibinfo{booktitle}{\emph{A networked self: Identity, community, and
  culture on social network sites}}.
\newblock \bibinfo{publisher}{Routledge}.
\newblock


\bibitem[Poria et~al\mbox{.}(2019)]%
        {poria2019emotion}
\bibfield{author}{\bibinfo{person}{Soujanya Poria}, \bibinfo{person}{Navonil
  Majumder}, \bibinfo{person}{Rada Mihalcea}, {and} \bibinfo{person}{Eduard
  Hovy}.} \bibinfo{year}{2019}\natexlab{}.
\newblock \showarticletitle{Emotion recognition in conversation: Research
  challenges, datasets, and recent advances}.
\newblock \bibinfo{journal}{\emph{IEEE access}}  \bibinfo{volume}{7}
  (\bibinfo{year}{2019}), \bibinfo{pages}{100943--100953}.
\newblock


\bibitem[Powers et~al\mbox{.}(2016)]%
        {powers2016treatment}
\bibfield{author}{\bibinfo{person}{Narelle Powers}, \bibinfo{person}{Judith
  Gullifer}, {and} \bibinfo{person}{Rhonda Shaw}.}
  \bibinfo{year}{2016}\natexlab{}.
\newblock \showarticletitle{When the treatment stops: A qualitative study of
  life post breast cancer treatment}.
\newblock \bibinfo{journal}{\emph{Journal of Health Psychology}}
  \bibinfo{volume}{21}, \bibinfo{number}{7} (\bibinfo{year}{2016}),
  \bibinfo{pages}{1371--1382}.
\newblock


\bibitem[Reddy et~al\mbox{.}(2024)]%
        {reddy2024audioinsight}
\bibfield{author}{\bibinfo{person}{Varun Reddy}, \bibinfo{person}{Zhiyuan
  Wang}, \bibinfo{person}{Emma~R Toner}, \bibinfo{person}{Maria~A Larrazabal},
  \bibinfo{person}{Mehdi Boukhechba}, \bibinfo{person}{Bethany~A Teachman},
  {and} \bibinfo{person}{Laura~E Barnes}.} \bibinfo{year}{2024}\natexlab{}.
\newblock \showarticletitle{Audioinsight: Detecting social contexts relevant to
  social anxiety from speech}. In \bibinfo{booktitle}{\emph{2024 12th
  International Conference on Affective Computing and Intelligent Interaction
  (ACII)}}. IEEE, \bibinfo{pages}{55--62}.
\newblock


\bibitem[Regulation(2018)]%
        {regulation2018general}
\bibfield{author}{\bibinfo{person}{Protection Regulation}.}
  \bibinfo{year}{2018}\natexlab{}.
\newblock \showarticletitle{General data protection regulation}.
\newblock \bibinfo{journal}{\emph{Intouch}}  \bibinfo{volume}{25}
  (\bibinfo{year}{2018}), \bibinfo{pages}{1--5}.
\newblock


\bibitem[Reimers(2019)]%
        {reimers2019sentence}
\bibfield{author}{\bibinfo{person}{N Reimers}.}
  \bibinfo{year}{2019}\natexlab{}.
\newblock \showarticletitle{Sentence-BERT: Sentence Embeddings using Siamese
  BERT-Networks}.
\newblock \bibinfo{journal}{\emph{arXiv preprint arXiv:1908.10084}}
  (\bibinfo{year}{2019}).
\newblock


\bibitem[Sadeghi et~al\mbox{.}(2023)]%
        {sadeghi2023exploring}
\bibfield{author}{\bibinfo{person}{Misha Sadeghi}, \bibinfo{person}{Bernhard
  Egger}, \bibinfo{person}{Reza Agahi}, \bibinfo{person}{Robert Richer},
  \bibinfo{person}{Klara Capito}, \bibinfo{person}{Lydia~Helene Rupp},
  \bibinfo{person}{Lena Schindler-Gmelch}, \bibinfo{person}{Matthias Berking},
  {and} \bibinfo{person}{Bjoern~M Eskofier}.} \bibinfo{year}{2023}\natexlab{}.
\newblock \showarticletitle{Exploring the capabilities of a language model-only
  approach for depression detection in text data}. In
  \bibinfo{booktitle}{\emph{2023 IEEE EMBS International Conference on
  Biomedical and Health Informatics (BHI)}}. IEEE, \bibinfo{pages}{1--5}.
\newblock


\bibitem[Salsman et~al\mbox{.}(2023)]%
        {salsman2023ehealth}
\bibfield{author}{\bibinfo{person}{John~M Salsman}, \bibinfo{person}{Laurie~E
  McLouth}, \bibinfo{person}{Janet~A Tooze}, \bibinfo{person}{Denisha
  Little-Greene}, \bibinfo{person}{Michael Cohn}, \bibinfo{person}{Mia~Sorkin
  Kehoe}, {and} \bibinfo{person}{Judith~T Moskowitz}.}
  \bibinfo{year}{2023}\natexlab{}.
\newblock \showarticletitle{An eHealth, positive emotion skills intervention
  for enhancing psychological well-being in young adult Cancer survivors:
  results from a multi-site, pilot feasibility trial}.
\newblock \bibinfo{journal}{\emph{International Journal of Behavioral
  Medicine}} \bibinfo{volume}{30}, \bibinfo{number}{5} (\bibinfo{year}{2023}),
  \bibinfo{pages}{639--650}.
\newblock


\bibitem[Salton and Buckley(1988)]%
        {salton1988term}
\bibfield{author}{\bibinfo{person}{Gerard Salton} {and}
  \bibinfo{person}{Christopher Buckley}.} \bibinfo{year}{1988}\natexlab{}.
\newblock \showarticletitle{Term-weighting approaches in automatic text
  retrieval}.
\newblock \bibinfo{journal}{\emph{Information processing \& management}}
  \bibinfo{volume}{24}, \bibinfo{number}{5} (\bibinfo{year}{1988}),
  \bibinfo{pages}{513--523}.
\newblock


\bibitem[Sanh et~al\mbox{.}(2019)]%
        {sanh2019distilbert}
\bibfield{author}{\bibinfo{person}{Victor Sanh}, \bibinfo{person}{Lysandre
  Debut}, \bibinfo{person}{Julien Chaumond}, {and} \bibinfo{person}{Thomas
  Wolf}.} \bibinfo{year}{2019}\natexlab{}.
\newblock \showarticletitle{DistilBERT, a distilled version of BERT: smaller,
  faster, cheaper and lighter}.
\newblock \bibinfo{journal}{\emph{arXiv preprint arXiv:1910.01108}}
  (\bibinfo{year}{2019}).
\newblock


\bibitem[Seabrook et~al\mbox{.}(2018)]%
        {seabrook2018predicting}
\bibfield{author}{\bibinfo{person}{Elizabeth~M Seabrook},
  \bibinfo{person}{Margaret~L Kern}, \bibinfo{person}{Ben~D Fulcher}, {and}
  \bibinfo{person}{Nikki~S Rickard}.} \bibinfo{year}{2018}\natexlab{}.
\newblock \showarticletitle{Predicting depression from language-based emotion
  dynamics: longitudinal analysis of Facebook and Twitter status updates}.
\newblock \bibinfo{journal}{\emph{Journal of medical Internet research}}
  \bibinfo{volume}{20}, \bibinfo{number}{5} (\bibinfo{year}{2018}),
  \bibinfo{pages}{e168}.
\newblock


\bibitem[Sharma et~al\mbox{.}(2023a)]%
        {sharma2023human}
\bibfield{author}{\bibinfo{person}{Ashish Sharma}, \bibinfo{person}{Inna~W
  Lin}, \bibinfo{person}{Adam~S Miner}, \bibinfo{person}{David~C Atkins}, {and}
  \bibinfo{person}{Tim Althoff}.} \bibinfo{year}{2023}\natexlab{a}.
\newblock \showarticletitle{Human--AI collaboration enables more empathic
  conversations in text-based peer-to-peer mental health support}.
\newblock \bibinfo{journal}{\emph{Nature Machine Intelligence}}
  \bibinfo{volume}{5}, \bibinfo{number}{1} (\bibinfo{year}{2023}),
  \bibinfo{pages}{46--57}.
\newblock


\bibitem[Sharma et~al\mbox{.}(2023b)]%
        {sharma2023cognitive}
\bibfield{author}{\bibinfo{person}{Ashish Sharma}, \bibinfo{person}{Kevin
  Rushton}, \bibinfo{person}{Inna~Wanyin Lin}, \bibinfo{person}{David Wadden},
  \bibinfo{person}{Khendra~G Lucas}, \bibinfo{person}{Adam~S Miner},
  \bibinfo{person}{Theresa Nguyen}, {and} \bibinfo{person}{Tim Althoff}.}
  \bibinfo{year}{2023}\natexlab{b}.
\newblock \showarticletitle{Cognitive reframing of negative thoughts through
  human-language model interaction}.
\newblock \bibinfo{journal}{\emph{arXiv preprint arXiv:2305.02466}}
  (\bibinfo{year}{2023}).
\newblock


\bibitem[Shi et~al\mbox{.}(2021)]%
        {10.1145/3448080}
\bibfield{author}{\bibinfo{person}{Dai Shi}, \bibinfo{person}{Dan Tao},
  \bibinfo{person}{Jiangtao Wang}, \bibinfo{person}{Muyan Yao},
  \bibinfo{person}{Zhibo Wang}, \bibinfo{person}{Houjin Chen}, {and}
  \bibinfo{person}{Sumi Helal}.} \bibinfo{year}{2021}\natexlab{}.
\newblock \showarticletitle{Fine-Grained and Context-Aware Behavioral
  Biometrics for Pattern Lock on Smartphones}.
\newblock \bibinfo{journal}{\emph{Proc. ACM Interact. Mob. Wearable Ubiquitous
  Technol.}} \bibinfo{volume}{5}, \bibinfo{number}{1}, Article
  \bibinfo{articleno}{33} (\bibinfo{date}{March} \bibinfo{year}{2021}),
  \bibinfo{numpages}{30}~pages.
\newblock
\href{https://doi.org/10.1145/3448080}{doi:\nolinkurl{10.1145/3448080}}


\bibitem[Shiffman et~al\mbox{.}(2008)]%
        {shiffman2008ecological}
\bibfield{author}{\bibinfo{person}{Saul Shiffman}, \bibinfo{person}{Arthur~A
  Stone}, {and} \bibinfo{person}{Michael~R Hufford}.}
  \bibinfo{year}{2008}\natexlab{}.
\newblock \showarticletitle{Ecological momentary assessment}.
\newblock \bibinfo{journal}{\emph{Annu. Rev. Clin. Psychol.}}
  \bibinfo{volume}{4}, \bibinfo{number}{1} (\bibinfo{year}{2008}),
  \bibinfo{pages}{1--32}.
\newblock


\bibitem[Sohn et~al\mbox{.}(2008)]%
        {sohn2008diary}
\bibfield{author}{\bibinfo{person}{Timothy Sohn}, \bibinfo{person}{Kevin~A Li},
  \bibinfo{person}{William~G Griswold}, {and} \bibinfo{person}{James~D
  Hollan}.} \bibinfo{year}{2008}\natexlab{}.
\newblock \showarticletitle{A diary study of mobile information needs}. In
  \bibinfo{booktitle}{\emph{Proceedings of the sigchi conference on human
  factors in computing systems}}. \bibinfo{pages}{433--442}.
\newblock


\bibitem[Stade et~al\mbox{.}(2024)]%
        {stade2024large}
\bibfield{author}{\bibinfo{person}{Elizabeth~C Stade},
  \bibinfo{person}{Shannon~Wiltsey Stirman}, \bibinfo{person}{Lyle~H Ungar},
  \bibinfo{person}{Cody~L Boland}, \bibinfo{person}{H~Andrew Schwartz},
  \bibinfo{person}{David~B Yaden}, \bibinfo{person}{Jo{\~a}o Sedoc},
  \bibinfo{person}{Robert~J DeRubeis}, \bibinfo{person}{Robb Willer}, {and}
  \bibinfo{person}{Johannes~C Eichstaedt}.} \bibinfo{year}{2024}\natexlab{}.
\newblock \showarticletitle{Large language models could change the future of
  behavioral healthcare: a proposal for responsible development and
  evaluation}.
\newblock \bibinfo{journal}{\emph{NPJ Mental Health Research}}
  \bibinfo{volume}{3}, \bibinfo{number}{1} (\bibinfo{year}{2024}),
  \bibinfo{pages}{12}.
\newblock


\bibitem[Stinson et~al\mbox{.}(2022)]%
        {stinson2022ecological}
\bibfield{author}{\bibinfo{person}{Lesleigh Stinson}, \bibinfo{person}{Yunchao
  Liu}, {and} \bibinfo{person}{Jesse Dallery}.}
  \bibinfo{year}{2022}\natexlab{}.
\newblock \showarticletitle{Ecological momentary assessment: a systematic
  review of validity research}.
\newblock \bibinfo{journal}{\emph{Perspectives on Behavior Science}}
  \bibinfo{volume}{45}, \bibinfo{number}{2} (\bibinfo{year}{2022}),
  \bibinfo{pages}{469--493}.
\newblock


\bibitem[Stone et~al\mbox{.}(2023)]%
        {stone2023evaluation}
\bibfield{author}{\bibinfo{person}{Arthur~A Stone}, \bibinfo{person}{Stefan
  Schneider}, {and} \bibinfo{person}{Joshua~M Smyth}.}
  \bibinfo{year}{2023}\natexlab{}.
\newblock \showarticletitle{Evaluation of pressing issues in ecological
  momentary assessment}.
\newblock \bibinfo{journal}{\emph{Annual Review of Clinical Psychology}}
  \bibinfo{volume}{19}, \bibinfo{number}{1} (\bibinfo{year}{2023}),
  \bibinfo{pages}{107--131}.
\newblock


\bibitem[Stone et~al\mbox{.}(2020)]%
        {stone2020evaluating}
\bibfield{author}{\bibinfo{person}{Arthur~A Stone}, \bibinfo{person}{Cheng
  K~Fred Wen}, \bibinfo{person}{Stefan Schneider}, {and}
  \bibinfo{person}{Doerte~U Junghaenel}.} \bibinfo{year}{2020}\natexlab{}.
\newblock \showarticletitle{Evaluating the effect of daily diary instructional
  phrases on respondents’ recall time frames: survey experiment}.
\newblock \bibinfo{journal}{\emph{Journal of Medical Internet Research}}
  \bibinfo{volume}{22}, \bibinfo{number}{2} (\bibinfo{year}{2020}),
  \bibinfo{pages}{e16105}.
\newblock


\bibitem[Straczkiewicz et~al\mbox{.}(2021)]%
        {straczkiewicz2021systematic}
\bibfield{author}{\bibinfo{person}{Marcin Straczkiewicz},
  \bibinfo{person}{Peter James}, {and} \bibinfo{person}{Jukka-Pekka Onnela}.}
  \bibinfo{year}{2021}\natexlab{}.
\newblock \showarticletitle{A systematic review of smartphone-based human
  activity recognition methods for health research}.
\newblock \bibinfo{journal}{\emph{NPJ Digital Medicine}} \bibinfo{volume}{4},
  \bibinfo{number}{1} (\bibinfo{year}{2021}), \bibinfo{pages}{148}.
\newblock


\bibitem[Tamir(2016)]%
        {tamir2016people}
\bibfield{author}{\bibinfo{person}{Maya Tamir}.}
  \bibinfo{year}{2016}\natexlab{}.
\newblock \showarticletitle{Why do people regulate their emotions? A taxonomy
  of motives in emotion regulation}.
\newblock \bibinfo{journal}{\emph{Personality and social psychology review}}
  \bibinfo{volume}{20}, \bibinfo{number}{3} (\bibinfo{year}{2016}),
  \bibinfo{pages}{199--222}.
\newblock


\bibitem[Teodorescu et~al\mbox{.}(2023)]%
        {teodorescu2023language}
\bibfield{author}{\bibinfo{person}{Daniela Teodorescu},
  \bibinfo{person}{Tiffany Cheng}, \bibinfo{person}{Alona Fyshe}, {and}
  \bibinfo{person}{Saif~M Mohammad}.} \bibinfo{year}{2023}\natexlab{}.
\newblock \showarticletitle{Language and mental health: Measures of emotion
  dynamics from text as linguistic biosocial markers}.
\newblock \bibinfo{journal}{\emph{arXiv preprint arXiv:2310.17369}}
  (\bibinfo{year}{2023}).
\newblock


\bibitem[Turcan and McKeown(2019)]%
        {turcan2019dreaddit}
\bibfield{author}{\bibinfo{person}{Elsbeth Turcan} {and}
  \bibinfo{person}{Kathleen McKeown}.} \bibinfo{year}{2019}\natexlab{}.
\newblock \showarticletitle{Dreaddit: A reddit dataset for stress analysis in
  social media}.
\newblock \bibinfo{journal}{\emph{arXiv preprint arXiv:1911.00133}}
  (\bibinfo{year}{2019}).
\newblock


\bibitem[Wang et~al\mbox{.}(2016)]%
        {wang2016dimensional}
\bibfield{author}{\bibinfo{person}{Jin Wang}, \bibinfo{person}{Liang-Chih Yu},
  \bibinfo{person}{K~Robert Lai}, {and} \bibinfo{person}{Xuejie Zhang}.}
  \bibinfo{year}{2016}\natexlab{}.
\newblock \showarticletitle{Dimensional sentiment analysis using a regional
  CNN-LSTM model}. In \bibinfo{booktitle}{\emph{Proceedings of the 54th annual
  meeting of the association for computational linguistics (volume 2: Short
  papers)}}. \bibinfo{pages}{225--230}.
\newblock


\bibitem[Wang et~al\mbox{.}(2023b)]%
        {wang2023power}
\bibfield{author}{\bibinfo{person}{Weichen Wang}, \bibinfo{person}{Weizhe Xu},
  \bibinfo{person}{Ayesha Chander}, \bibinfo{person}{Subigya Nepal},
  \bibinfo{person}{Benjamin Buck}, \bibinfo{person}{Serguei Pakhomov},
  \bibinfo{person}{Trevor Cohen}, \bibinfo{person}{Dror Ben-Zeev}, {and}
  \bibinfo{person}{Andrew Campbell}.} \bibinfo{year}{2023}\natexlab{b}.
\newblock \showarticletitle{The Power of Speech in the Wild: Discriminative
  Power of Daily Voice Diaries in Understanding Auditory Verbal Hallucinations
  Using Deep Learning}.
\newblock \bibinfo{journal}{\emph{Proceedings of the ACM on Interactive,
  Mobile, Wearable and Ubiquitous Technologies}} \bibinfo{volume}{7},
  \bibinfo{number}{3} (\bibinfo{year}{2023}), \bibinfo{pages}{1--29}.
\newblock


\bibitem[Wang et~al\mbox{.}(2023a)]%
        {wang2023detecting}
\bibfield{author}{\bibinfo{person}{Zhiyuan Wang}, \bibinfo{person}{Maria~A
  Larrazabal}, \bibinfo{person}{Mark Rucker}, \bibinfo{person}{Emma~R Toner},
  \bibinfo{person}{Katharine~E Daniel}, \bibinfo{person}{Shashwat Kumar},
  \bibinfo{person}{Mehdi Boukhechba}, \bibinfo{person}{Bethany~A Teachman},
  {and} \bibinfo{person}{Laura~E Barnes}.} \bibinfo{year}{2023}\natexlab{a}.
\newblock \showarticletitle{Detecting social contexts from mobile sensing
  indicators in virtual interactions with socially anxious individuals}.
\newblock \bibinfo{journal}{\emph{Proceedings of the ACM on interactive,
  mobile, wearable and ubiquitous technologies}} \bibinfo{volume}{7},
  \bibinfo{number}{3} (\bibinfo{year}{2023}), \bibinfo{pages}{1--26}.
\newblock


\bibitem[Wang et~al\mbox{.}(2024a)]%
        {wang2024rapport}
\bibfield{author}{\bibinfo{person}{Zhiyuan Wang}, \bibinfo{person}{Varun
  Reddy}, \bibinfo{person}{Karen Ingersoll}, \bibinfo{person}{Tabor
  Flickinger}, {and} \bibinfo{person}{Laura~E Barnes}.}
  \bibinfo{year}{2024}\natexlab{a}.
\newblock \showarticletitle{Rapport Matters: Enhancing HIV mHealth
  Communication through Linguistic Analysis and Large Language Models}. In
  \bibinfo{booktitle}{\emph{Extended Abstracts of the CHI Conference on Human
  Factors in Computing Systems}}. \bibinfo{pages}{1--8}.
\newblock


\bibitem[Wang et~al\mbox{.}(2024b)]%
        {wang2024pallm}
\bibfield{author}{\bibinfo{person}{Zhiyuan Wang}, \bibinfo{person}{Fangxu
  Yuan}, \bibinfo{person}{Virginia LeBaron}, \bibinfo{person}{Tabor
  Flickinger}, {and} \bibinfo{person}{Laura~E Barnes}.}
  \bibinfo{year}{2024}\natexlab{b}.
\newblock \showarticletitle{PALLM: Evaluating and Enhancing PALLiative Care
  Conversations with Large Language Models}.
\newblock \bibinfo{journal}{\emph{ACM Transactions on Computing for
  Healthcare}} (\bibinfo{year}{2024}).
\newblock


\bibitem[Wu et~al\mbox{.}(2024)]%
        {wu2024mindshift}
\bibfield{author}{\bibinfo{person}{Ruolan Wu}, \bibinfo{person}{Chun Yu},
  \bibinfo{person}{Xiaole Pan}, \bibinfo{person}{Yujia Liu},
  \bibinfo{person}{Ningning Zhang}, \bibinfo{person}{Yue Fu},
  \bibinfo{person}{Yuhan Wang}, \bibinfo{person}{Zhi Zheng},
  \bibinfo{person}{Li Chen}, \bibinfo{person}{Qiaolei Jiang}, {et~al\mbox{.}}}
  \bibinfo{year}{2024}\natexlab{}.
\newblock \showarticletitle{MindShift: Leveraging Large Language Models for
  Mental-States-Based Problematic Smartphone Use Intervention}. In
  \bibinfo{booktitle}{\emph{Proceedings of the CHI Conference on Human Factors
  in Computing Systems}}. \bibinfo{pages}{1--24}.
\newblock


\bibitem[Xiong et~al\mbox{.}(2024)]%
        {xiong2024converging}
\bibfield{author}{\bibinfo{person}{Haoyi Xiong}, \bibinfo{person}{Zhiyuan
  Wang}, \bibinfo{person}{Xuhong Li}, \bibinfo{person}{Jiang Bian},
  \bibinfo{person}{Zeke Xie}, \bibinfo{person}{Shahid Mumtaz},
  \bibinfo{person}{Anwer Al-Dulaimi}, {and} \bibinfo{person}{Laura~E Barnes}.}
  \bibinfo{year}{2024}\natexlab{}.
\newblock \showarticletitle{Converging paradigms: The synergy of symbolic and
  connectionist ai in llm-empowered autonomous agents}.
\newblock \bibinfo{journal}{\emph{arXiv preprint arXiv:2407.08516}}
  (\bibinfo{year}{2024}).
\newblock


\bibitem[Xu et~al\mbox{.}(2024)]%
        {xu2024mental}
\bibfield{author}{\bibinfo{person}{Xuhai Xu}, \bibinfo{person}{Bingsheng Yao},
  \bibinfo{person}{Yuanzhe Dong}, \bibinfo{person}{Saadia Gabriel},
  \bibinfo{person}{Hong Yu}, \bibinfo{person}{James Hendler},
  \bibinfo{person}{Marzyeh Ghassemi}, \bibinfo{person}{Anind~K Dey}, {and}
  \bibinfo{person}{Dakuo Wang}.} \bibinfo{year}{2024}\natexlab{}.
\newblock \showarticletitle{Mental-llm: Leveraging large language models for
  mental health prediction via online text data}.
\newblock \bibinfo{journal}{\emph{Proceedings of the ACM on Interactive,
  Mobile, Wearable and Ubiquitous Technologies}} \bibinfo{volume}{8},
  \bibinfo{number}{1} (\bibinfo{year}{2024}), \bibinfo{pages}{1--32}.
\newblock


\bibitem[Yang et~al\mbox{.}(2023)]%
        {yang2023towards}
\bibfield{author}{\bibinfo{person}{Kailai Yang}, \bibinfo{person}{Shaoxiong
  Ji}, \bibinfo{person}{Tianlin Zhang}, \bibinfo{person}{Qianqian Xie},
  \bibinfo{person}{Ziyan Kuang}, {and} \bibinfo{person}{Sophia Ananiadou}.}
  \bibinfo{year}{2023}\natexlab{}.
\newblock \showarticletitle{Towards interpretable mental health analysis with
  large language models}.
\newblock \bibinfo{journal}{\emph{arXiv preprint arXiv:2304.03347}}
  (\bibinfo{year}{2023}).
\newblock


\bibitem[Yang et~al\mbox{.}(2022)]%
        {yang2022large}
\bibfield{author}{\bibinfo{person}{Xi Yang}, \bibinfo{person}{Aokun Chen},
  \bibinfo{person}{Nima PourNejatian}, \bibinfo{person}{Hoo~Chang Shin},
  \bibinfo{person}{Kaleb~E Smith}, \bibinfo{person}{Christopher Parisien},
  \bibinfo{person}{Colin Compas}, \bibinfo{person}{Cheryl Martin},
  \bibinfo{person}{Anthony~B Costa}, \bibinfo{person}{Mona~G Flores},
  {et~al\mbox{.}}} \bibinfo{year}{2022}\natexlab{}.
\newblock \showarticletitle{A large language model for electronic health
  records}.
\newblock \bibinfo{journal}{\emph{NPJ digital medicine}} \bibinfo{volume}{5},
  \bibinfo{number}{1} (\bibinfo{year}{2022}), \bibinfo{pages}{194}.
\newblock


\bibitem[Yang(2019)]%
        {yang2019xlnet}
\bibfield{author}{\bibinfo{person}{Zhilin Yang}.}
  \bibinfo{year}{2019}\natexlab{}.
\newblock \showarticletitle{XLNet: Generalized Autoregressive Pretraining for
  Language Understanding}.
\newblock \bibinfo{journal}{\emph{arXiv preprint arXiv:1906.08237}}
  (\bibinfo{year}{2019}).
\newblock


\bibitem[Yao et~al\mbox{.}(2024)]%
        {yao2024personalised}
\bibfield{author}{\bibinfo{person}{Tsungcheng Yao}, \bibinfo{person}{Ernest
  Foo}, {and} \bibinfo{person}{Sebastian Binnewies}.}
  \bibinfo{year}{2024}\natexlab{}.
\newblock \showarticletitle{Personalised Abusive Language Detection Using LLMs
  and Retrieval-Augmented Generation}. In \bibinfo{booktitle}{\emph{Proceedings
  of the 7th International Conference on Natural Language and Speech Processing
  (ICNLSP 2024)}}. \bibinfo{pages}{92--98}.
\newblock


\bibitem[Zabora et~al\mbox{.}(2001)]%
        {zabora2001prevalence}
\bibfield{author}{\bibinfo{person}{James Zabora}, \bibinfo{person}{Karlynn
  BrintzenhofeSzoc}, \bibinfo{person}{Barbara Curbow}, \bibinfo{person}{Craig
  Hooker}, {and} \bibinfo{person}{Steven Piantadosi}.}
  \bibinfo{year}{2001}\natexlab{}.
\newblock \showarticletitle{The prevalence of psychological distress by cancer
  site}.
\newblock \bibinfo{journal}{\emph{Psycho-oncology: journal of the
  psychological, social and behavioral dimensions of Cancer}}
  \bibinfo{volume}{10}, \bibinfo{number}{1} (\bibinfo{year}{2001}),
  \bibinfo{pages}{19--28}.
\newblock


\bibitem[Zebrack(2000)]%
        {zebrack2000cancer}
\bibfield{author}{\bibinfo{person}{Brad~J Zebrack}.}
  \bibinfo{year}{2000}\natexlab{}.
\newblock \showarticletitle{Cancer survivor identity and quality of life}.
\newblock \bibinfo{journal}{\emph{Cancer practice}} \bibinfo{volume}{8},
  \bibinfo{number}{5} (\bibinfo{year}{2000}), \bibinfo{pages}{238--242}.
\newblock


\bibitem[Zeng et~al\mbox{.}(2006)]%
        {zeng2006audio}
\bibfield{author}{\bibinfo{person}{Zhihong Zeng}, \bibinfo{person}{Yuxiao Hu},
  \bibinfo{person}{Yun Fu}, \bibinfo{person}{Thomas~S Huang},
  \bibinfo{person}{Glenn~I Roisman}, {and} \bibinfo{person}{Zhen Wen}.}
  \bibinfo{year}{2006}\natexlab{}.
\newblock \showarticletitle{Audio-visual emotion recognition in adult
  attachment interview}. In \bibinfo{booktitle}{\emph{Proceedings of the 8th
  international conference on Multimodal interfaces}}.
  \bibinfo{pages}{139--145}.
\newblock


\bibitem[Zhao et~al\mbox{.}(2024)]%
        {zhao2024retrieval}
\bibfield{author}{\bibinfo{person}{Penghao Zhao}, \bibinfo{person}{Hailin
  Zhang}, \bibinfo{person}{Qinhan Yu}, \bibinfo{person}{Zhengren Wang},
  \bibinfo{person}{Yunteng Geng}, \bibinfo{person}{Fangcheng Fu},
  \bibinfo{person}{Ling Yang}, \bibinfo{person}{Wentao Zhang}, {and}
  \bibinfo{person}{Bin Cui}.} \bibinfo{year}{2024}\natexlab{}.
\newblock \showarticletitle{Retrieval-augmented generation for ai-generated
  content: A survey}.
\newblock \bibinfo{journal}{\emph{arXiv preprint arXiv:2402.19473}}
  (\bibinfo{year}{2024}).
\newblock


\end{thebibliography}

\appendix 

\clearpage 
\section{Appendix}

\subsection{Prompt Design}\label{apd:prompt}

\begin{wrapfigure}{r}{0.385\linewidth}
    \centering
    \includegraphics[width=\linewidth]{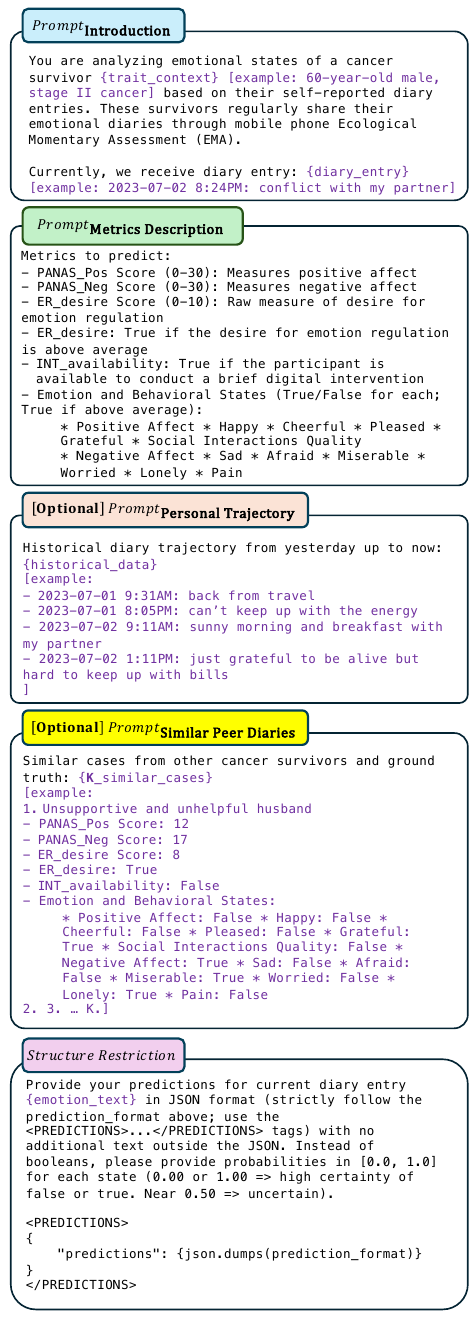}
    \caption{Detailed CALLM prompt structure.}
    \label{fig:prompts}
\end{wrapfigure}

Our prompt design consists of five key components, as shown in Figure~\ref{fig:prompts}, which extends the simplified demonstration in Figure~\ref{fig:prompt_demo} with comprehensive prompt language details:

\begin{itemize}
    \item \textbf{Introduction}: Provides essential context about the cancer survivor being analyzed:
        \begin{itemize}
            \item Trait information extracted from participant profiles (e.g., age, gender, cancer type and stage)
            \item Data collection context (mobile EMA-based diary entries)
            \item Current diary entry timestamp and content
        \end{itemize}
    
    \item \textbf{Metrics Description}: Defines the target metrics to predict:
        \begin{itemize}
            \item PANAS scores for positive and negative affect (0-30 scale)
            \item Emotion regulation desire score (0-10 scale) and its binary indicator derived from individual mean comparison
            \item Intervention availability (binary)
            \item Binary states for specific emotions and behaviors derived from individual mean comparisons (binary)
        \end{itemize}
    
    \item \textbf{[Optional] Personal Trajectory}: Includes historical diary entries from the previous day up to the current moment:
        \begin{itemize}
            \item Retrieves timestamps and contents of recent diary entries within the time range identified (e.g., since last or current day before the current diary)
            \item Ordered chronologically to show temporal progression
        \end{itemize}
    
    \item \textbf{[Optional] Similar Peer Diaries}: Presents comparable cases retrieved through RAG:
        \begin{itemize}
            \item Top-K semantically similar diary entries selected using FAISS vector similarity search, serving as few-shot learning examples
            \item Each case includes the complete diary text and all associated ground truth
        \end{itemize}
    
    \item \textbf{Structure Restriction}: Enforces strict JSON formatting for predictions:
        \begin{itemize}
            \item Requires probability scores in [0.0, 1.0] range for all binary states, with 0.5 specified as uncertain.
            \item Mandates $<PREDICTIONS>$ tags to ensure reliable parsing and post-processing
        \end{itemize}
\end{itemize}

This structured prompt design ensures consistent input format while providing comprehensive context for emotional state prediction. The modular design allows for flexible configuration of optional components based on data availability and specific prediction needs.

\section{Baseline Hyperparameter Tuning} \label{appendix:hyper-parameter}
We conducted extensive hyperparameter tuning for all baseline models to ensure a fair comparison with the CALLM framework. The tuning process involved nested cross-validation to avoid data leakage, with parameters optimized for the multitask learning objective across all target variables. In our nested scheme, the outer loop consisted of the 5-fold grouped cross-validation (stratified by participant), providing unbiased performance estimates, while the inner loop performed hyperparameter selection using 5-fold cross-validation on the training portion only. This nested approach ensured that hyperparameter selection had no access to the test data, maintaining the integrity of our reported performance metrics.

For traditional machine learning approaches, we used Logistic Regression classifiers with various text representation methods. For Bag-of-Words (BoW) and TF-IDF feature extraction, we explored a comprehensive parameter space. Vectorization parameters included various document frequency thresholds (min\_df values of 1, 2, and 5; max\_df values of 0.9, 0.95, and 1.0) and n-gram ranges (unigrams only, and unigrams with bigrams). For TF-IDF specifically, we additionally tuned normalization methods (L1 and L2) and tested both with and without inverse document frequency weighting. The Logistic Regression classifier hyperparameters encompassed regularization strengths (C values of 0.1, 1.0, and 10.0), penalty types (L1 and L2), solver method (liblinear), and class weighting approaches (balanced versus none). This multi-output configuration of Logistic Regression classifiers enabled simultaneous prediction of all emotional and behavioral states.

For transformer-based models (BERT, RoBERTa, EmoBERT, etc.), we tuned several critical parameters affecting both computational efficiency and model performance. These included batch sizes (8, 16, and 32), learning rates (1e-5, 5e-5, 1e-4, and 5e-4, 1e-3), maximum sequence lengths (128 and 256), and training epochs (2, 3, 4, and 6). To mitigate overfitting, we implemented multiple regularization strategies: classifier dropout rates (0.1, 0.2, and 0.3), weight decay (0.01 and 0.1), and architecture-specific dropouts for attention mechanisms and hidden states (0.1 and 0.2). We also employed early stopping with a patience value of 2, halting training when validation performance ceased to improve, which both prevented overfitting and reduced computational costs. Due to computational constraints, we employed a two-stage tuning approach for transformer models. First, we conducted a coarse grid search on a subset of hyperparameter combinations using groups 1-3 for training and groups 4-5 for validation. Then, we fine-tuned the most promising configurations on the full dataset. For memory-intensive models like SentenceBERT and Emotion-Transformer, we used smaller batch sizes (8, 16) and implemented chunked processing to manage memory efficiently.

All tuning was performed by optimizing the mean balanced accuracy across all four target variables (positive affect, negative affect, emotion regulation desire, and intervention availability). The best parameters for each model were selected based on validation performance and then applied consistently across all five stratified groups of the final evaluation.

\end{document}